\newif\ifAGU

\ifAGU
\documentclass[draft]{agujournal2019}
\usepackage{url} 
\usepackage{lineno}
\usepackage[inline]{trackchanges} 
\usepackage{soul}
\linenumbers

\usepackage{amsfonts}       

\usepackage{bm}
\usepackage[version=4]{mhchem}

\draftfalse

\journalname{Journal of Advances in Modeling Earth Systems (JAMES)}

\else

\documentclass[11pt]{article}
\usepackage{graphicx} 
\usepackage[a4paper, 
left=30mm,
bottom=20mm,
 top=20mm,
 right=15mm]{geometry}
 \usepackage{gensymb}
 \usepackage{amsmath}
 \usepackage{amssymb}

\usepackage{xcolor}
\usepackage{booktabs}
\usepackage{url} 

\usepackage{bm}
\usepackage[version=4]{mhchem}
\usepackage{ragged2e}
\usepackage{soul}

\usepackage{helvet}
\usepackage{titlesec}
\usepackage{xcolor}
\titleformat*{\section}{\LARGE\bfseries}
\titleformat*{\subsection}{\Large\bfseries}
\titleformat*{\subsubsection}{\large\bfseries}
\titleformat*{\paragraph}{\large\bfseries}

\usepackage[style=numeric-comp, url=false,isbn=false,doi=true, eprint=false,giveninits=true,maxbibnames=99,maxcitenames=2,date=year,sorting=none,sortcites=true, natbib=true]{biblatex}
\let\citeA\citet

\def\title#1{
{\centering \Huge\bf #1 \vskip14pt}
\def\thetitle{#1}}

\def\authors#1{{\centering \normalsize\bf #1\vskip12pt}}
\def\affil#1{$^{#1}$\ignorespaces}
\def\affiliation#1#2{\vskip-.5\parskip\relax{\centering{\footnotesize
$^{#1}$#2\relax}\vskip-\parskip}}

\def\correspondingauthor#1#2{{{\noindent\vrule
height 18pt width0pt\relax\hbox to-8pt{}{\small
Corresponding author: #1,
{\tt #2}}}}}

\def\thanks#1{{\renewcommand\thefootnote{\@fnsymbol\c@footnote}%
    \def\@makefnmark{\rlap{\@textsuperscript{\normalfont\@thefnmark}}}%
    \long\def\@makefntext##1{\noindent\hskip-9pt\hb@xt@1.8em{%
                \hss\@textsuperscript{\normalfont\@thefnmark}}##1}\footnote{#1}\
		}}

\def\keypoints{\vskip24pt\vskip1sp\subsection*{Key Points:}
\begin{itemize}}
\def\endkeypoints{\end{itemize}}

\def\abstract{{{\section*{Abstract}}}\vskip-\parskip}

\def\acknowledgments{\vskip12pt\noindent{\bf
Acknowledgments\vrule depth 6pt
width0pt\relax}\\*\noindent\ignorespaces}

\fi 

\usepackage{booktabs}
\usepackage{tabularx}
\usepackage{pifont}
\newcommand{\cmark}{\ding{51}}%
\newcommand{\xmark}{\ding{55}}%

\begin{document}

\title{Atmospheric Transport Modeling of \ce{CO2} with Neural Networks} 

%
%




\authors{Vitus Benson\affil{1,2,3}, Ana Bastos\affil{4,2}, Christian Reimers\affil{1,2}, Alexander~J.~Winkler\affil{1,2}, Fanny~Yang\affil{3}, and Markus Reichstein\affil{1,2}}

\affiliation{1}{Max Planck Institute for Biogeochemistry, Jena, Germany}
\affiliation{2}{ELLIS Unit Jena, Jena, Germany}
\affiliation{3}{ETH Zürich, Zürich, Switzerland}
\affiliation{4}{Leipzig University, Leipzig, Germany}




\correspondingauthor{Vitus Benson}{vbenson@bgc-jena.mpg.de}



\begin{keypoints}
\item CarbonBench: a systematic benchmark for machine learning emulators of atmospheric tracer transport
\item Adapted SwinTransformer deep neural network to achieve stable and mass-conserving transport of \ce{CO2} by including physical constraints
\item UNet, GraphCast and Spherical Fourier Neural Operator baselines with the same customization are also strong models, for shorter lead times (up to 90 days)
\end{keypoints}

%
%

%
%


\begin{abstract} 
Accurately describing the distribution of \ce{CO2} in the atmosphere with atmospheric tracer transport models is essential for greenhouse gas monitoring and verification support systems to aid implementation of international climate agreements. Large deep neural networks are poised to revolutionize weather prediction, which requires 3D modeling of the atmosphere. While similar in this regard, atmospheric transport modeling is subject to new challenges. Both, stable predictions for longer time horizons and mass conservation throughout need to be achieved, while IO plays a larger role compared to computational costs. In this study we explore four different deep neural networks (UNet, GraphCast, Spherical Fourier Neural Operator and SwinTransformer) which have proven as state-of-the-art in weather prediction to assess their usefulness for atmospheric tracer transport modeling. For this, we assemble the CarbonBench dataset, a systematic benchmark tailored for machine learning emulators of Eulerian atmospheric transport. Through architectural adjustments, we decouple the performance of our emulators from the distribution shift caused by a steady rise in atmospheric \ce{CO2}. More specifically, we center \ce{CO2} input fields to zero mean and then use an explicit flux scheme and a mass fixer to assure mass balance. This design enables stable and mass conserving transport for over 6 months with all four neural network architectures. In our study, the SwinTransformer displays particularly strong emulation skill (90-day $R^2 > 0.99$), with physically plausible emulation even for forward runs of multiple years. This work paves the way forward towards high resolution forward and inverse modeling of inert trace gases with neural networks.
\end{abstract}
\justifying

\ifAGU
\else
\newpage
\fi

\section{Introduction}
Limiting greenhouse gas emissions in line with the Paris agreement to mitigate anthropogenic climate change requires monitoring, reporting and verification (MRV) efforts, especially of carbon dioxide (\ce{CO2}) \cite{friedlingstein.etal_2023}. Atmospheric measurements of \ce{CO2} from ground-based observatories, aircraft and satellite can provide independent, science-based estimates. However, these observations represent the concentration in the free air, not directly the emissions and other surface fluxes. Atmospheric transport models build the necessary bridge, allowing to understand \ce{CO2} concentrations from the perspective of anthropogenic emissions, biosphere and ocean fluxes \cite{kaminski.heimann_2001,gurney.etal_2002a, ciais.etal_2011}. They solve the continuity equation of the mass of \ce{CO2} in the atmosphere by computing horizontal advection and vertical movement of air parcels using driving meteorological reanalysis fields \cite{brasseur.jacob_2017}.

Since its early ages in the late 1980s, solving 3D tracer transport with numerical schemes has been hampered by prohibitive computational costs when going to higher resolution \cite{williamson_1992}. Yet, low resolution transport models, suffer from a variety of modeling errors \cite{schuh.etal_2019a, gaubert.etal_2019a}. More specifically, representations of convective transport \cite{belikov.etal_2013,  munassar.etal_2023,remaud.etal_2023a, schuh.jacobson_2023}, turbulent vertical mixing \cite{kretschmer.etal_2012}, summertime diabatic mixing \cite{jin.etal_2024}, numerical advection scheme \cite{agusti-panareda.etal_2017, eastham.jacob_2017} and reanalysed meteorological fields \cite{yu.etal_2018, zhang.etal_2021b} in atmospheric transport models display significant uncertainties. Increasing resolution has been proposed as one potential remedy to the situation \cite{remaud.etal_2018, agusti-panareda.etal_2019}.

However, a primary application of transport models is in inverse modeling of the surface fluxes to contribute regularly to MRV efforts such as the annual Global Carbon Budget updates \cite{friedlingstein.etal_2023}. Starting from prior surface fluxes, the transport model is used to map them to atmospheric concentrations which can be compared against observations to subsequently optimize the fluxes through Bayesian calibration \cite{rodenbeck.etal_2003, chevallier.etal_2005, rodenbeck_2005, chevallier.etal_2006, peters.etal_2007, vanderlaan-luijkx.etal_2017, remaud.etal_2018, rodenbeck.etal_2018, chandra.etal_2022}). This iterative process typically requires many expensive calls of the transport model and its adjoint, thereby rendering the usage of high fidelity solvers difficult \cite{chevallier.etal_2023}.

Recently, AI-based emulation has revolutionized numerical weather prediction: deep neural networks trained on high resolution meteorological reanalysis can both, outpace and outperform, traditional medium-range weather forecasting systems \cite{bi.etal_2022, keisler_2022, pathak.etal_2022, bonev.etal_2023, chen.etal_2023, kochkov.etal_2023, lam.etal_2023a, price.etal_2023}. Crucially, these emulators require less vertical layers, allow for larger time steps and leverage computing infrastructure optimized for matrix multiplication like GPUs. Hence, the neural networks learn to solve the Navier Stokes equations, by implicitly representing both, the large-scale dynamics that could be explicitly solved, and subgrid-scale processes that have to be parameterized, some works even make this division explicit \cite{krasnopolsky.fox-rabinovitz_2006, arcomano.etal_2022, kochkov.etal_2023}. Furthermore, foundation models are being introduced which support other tasks beyond medium-range weather forecasting, such as climate modeling \cite{nguyen.etal_2023a, lessig.etal_2023} or short-term forecasts of atmospheric composition \cite{bodnar.etal_2024a}.

Modeling the atmospheric carbon cycle with neural networks has not yet gathered as much attention. Still, there are works on emulating the footprints obtained from Lagrangian particle dispersion models of \ce{CH4}, which are useful for regional inverse modeling: Over a few UK regions, the NAME model has been emulated with CNNs \cite{cartwright.etal_2023} and with Gradient Boosting Trees  \cite{fillola.etal_2022} and over a few US regions, STILT has been emulated with FootNet \cite{he.etal_2023}, also a CNN. If more broadly considering approaches to modeling the \ce{CO2} and \ce{CH4} surface fluxes, machine learning has been used to upscale eddy covariance measurements as functions of climate and remote sensing to the globe, to obtain land fluxes of \ce{CH4} \cite{mcnicol.etal_2023a} and \ce{CO2} \cite{jung.etal_2011, tramontana.etal_2016, jung.etal_2020, nelson.etal_2024}. For the latter, \citeA{upton.etal_2024} recently introduced additional atmospheric constraints, bridging between atmospheric inverse modeling and machine learning-based upscaling.

Here, we introduce atmospheric transport modeling of \ce{CO2} with neural network emulators. Our main contributions are three-fold:
\begin{enumerate}
    \item We create a new dataset (CarbonBench), the first systematic benchmark for training and testing machine learning emulators of Eulerian atmospheric transport.
    \item We develop a SwinTransformer-based emulator tailored for transport modeling through physics-based adjustments that allow for strong empirical performance: forward runs with global RMSE below 1 ppm are possible for multiple years.
    \item We compare performance against three other large deep neural network architectures (UNet, GraphCast \& SFNO). While the SwinTransformer outperforms, with our generic architectural changes also the baselines achieve stable and mass-conserving transport for over 6 months.
\end{enumerate}
Thus, we provide the first step towards a high resolution \ce{CO2} inversion system leveraging AI to support the World Meteorological Organizations Global Greenhouse Gas Watch (G3W) and other efforts in line with the Paris agreement.

\section{Methods}

\begin{figure}
    \centering
    \includegraphics[width = \textwidth]{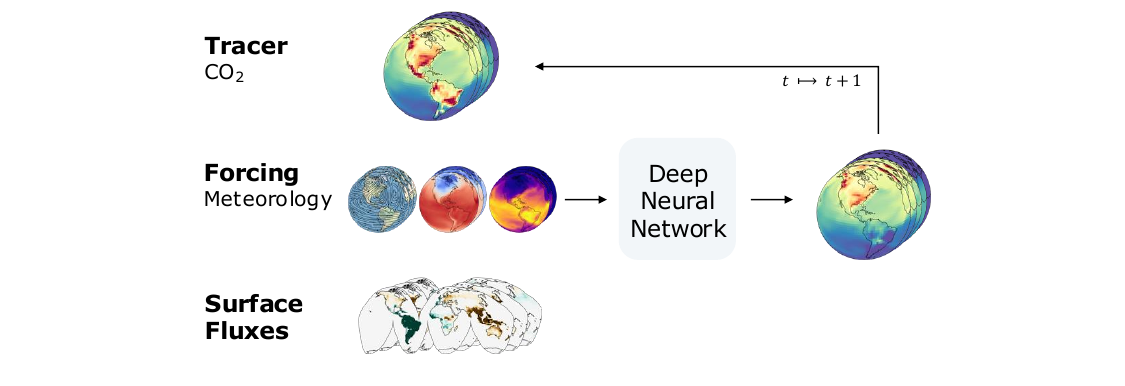}
    \caption{Offline atmospheric tracer transport modeling with deep neural networks.}
    \label{fig:concept}
\end{figure}

\subsection{Task}
In this work, we are tackling offline tracer transport with neural networks. That is, we solve the continuity equation for the inert trace gas \ce{CO2} given prescribed meteorology. In other words, we predict the 3D field of \ce{CO2} concentration in the atmosphere at time $t+1$ given the \ce{CO2} concentration field from the previous time step $t$ and meteorology and surface fluxes as additional inputs (fig.~\ref{fig:concept}). Like conventional solvers, our learned neural networks are autoregressive: longer forward runs can be produced by feeding the predicted \ce{CO2} concentrations back in as inputs, alongside prescribed fluxes and meteorology from the next time step. This allows in principle to generate arbitrarily long trajectories of \ce{CO2} fields, if sufficient forcing data is available.

More specifically consider the \ce{CO2} mass mixing ratio $\mu$, a source/sink term $\Sigma$ and the vector of wind fields $V$, then tracer transport follows from integrating
\begin{equation}\label{eq:continuity}
    \frac{d\mu}{dt} + V \cdot \nabla \mu = \Sigma
\end{equation}
over the spherical shell $\mathcal{D} = S^2 \times [r, r+h] \subset \mathbb{R}^3$, with $S^2$ the sphere, $r$ the radius of Earth and $h$ the height of the atmosphere. The integration is typically done by specifying von Neumann boundary conditions $\frac{d\mu}{d\bm{n}} = 0$, with $n$ being the outward-facing normal derivative on $D$, in other words: the flux out of the atmosphere is none. This would model surface fluxes with the source/sink term $\Sigma$, allowing for emissions inside the atmosphere. However, one may alternatively want to model surface fluxes as the lower boundary condition. In offline tracer transport models, the winds $V$ are prescribed. An alternative approach would be online tracer transport, where in addition to the tracer transport, the full atmospheric dynamics are modeled\cite{patra.etal_2018}. 

When numerically integrating the continuity equation, one needs to discretize over a grid, which requires splitting the operator into resolved and unresolved scales. For atmospheric transport, one furthermore typically splits the operator into horizontal advection and vertical convection, whereby for the former any subgrid-scale closure is ignored, but for the latter it is parameterized \cite{heimann.korner_2003}. Hence we end up with the equation
\begin{equation}
    \frac{d\mu}{dt} + u \frac{d\mu}{dx} + v \frac{d\mu}{dy} + w(\omega, T, q, z) \frac{d\mu}{dz} = \Sigma
\end{equation}
with the vertical velocity $w$ being a function of updraft $\omega$, temperature $T$, specific humidity $q$ and geopotential height $z$. Throughout this work, we use neural networks to solve directly for the time derivative:
\begin{equation}\label{eq:emulation}
    \frac{d\mu}{dt} = f(\mu, u, v, \omega, T, q, z, \ldots ; \theta)
\end{equation}
with $f(\cdot ; \theta)$ being a neural networks with parameters $\theta$. We then integrate using Euler steps $\mu_{t+1} = \mu_{t} + \frac{d\mu}{dt}$. During training, this means we approximate $\Delta\mu_t = \mu_{t+1} - \mu_t$ with the neural network $f(\cdot ; \theta)$ by optimizing parameters through minimizing the squared loss:
\begin{equation}
    \hat \theta = \operatorname*{arg\,min}_\theta \mathbb{E} ||(f(X_t; \theta) - \Delta\mu_t)||_2^2
\end{equation}

\subsection{CarbonBench Dataset}
For training the neural network emulators, we collect two existing datasets and reprocess them into a deep learning-ready format. The first dataset (CarbonTracker) is an inversion of \ce{CO2}, i.e. it has been obtained by optimizing the surface fluxes by transporting them and then matching modeled atmospheric concentrations against observed ones. The second dataset (ObsPack) contains atmospheric measurements of \ce{CO2}, allowing to compare our model predictions against an absolute baseline, independent of the training targets.

\subsubsection{CarbonTracker}
The CarbonTracker North America inversions \cite{peters.etal_2007} utilize the TM5 \cite{krol.etal_2005} transport model and the ensemble Kalman filter to perform inverse modeling of the surface fluxes. More specifically, they start with a set of prior fluxes for the land and ocean (e.g. from Earth system models) and add these to prescribed fluxes for anthropogenic emissions and wildfires to obtain a first version of total \ce{CO2} surface fluxes. In a next step, they leverage an atmospheric transport model and the ensemble Kalman filter to optimize the surface fluxes such that they match well to observed data of atmospheric \ce{CO2} concentrations. Finally, the optimized fluxes are transported one more time to obtain a 3D field of atmospheric \ce{CO2} concentrations. Here, we only use the final product from the inverse modeling process: the optimized surface fluxes and corresponding 3D fields. Moreover, we treat all surface fluxes as prescribed inputs, and not just the anthropogenic and wildfire components.

We collect 3D atmospheric \ce{CO2} concentration fields, 2D \ce{CO2} surface fluxes and 3D meteorological fields of $q, T, u, v, \omega, z$ \ce from the CarbonTracker CT2022 version \cite{jacobson.etal_2023a}. These represent a closed system, i.e. they fulfill a discretized version of the continuity eq.~\ref{eq:continuity}. Moreover, as they have been produced through inverse modeling, they are also closely resembling observations of atmospheric \ce{CO2}.

We prepare three versions of the dataset through aggregation that allow for quicker experimentation and testing of methods at multiple resolution. Each dataset we  split into training (years 2000-2016), validation (2017) and testing (2018-2020) sets, the three resolutions are: 
\begin{itemize}
    \item LowRes: $5.625^{\circ} \times 5.625^{\circ} \times 10$ hybrid vertical levels $\times 6$h.
    \item MidRes: $2.8125^{\circ} \times 2.8125^{\circ} \times 20$ hybrid vertical levels $\times 6$h.
    \item OrigRes: $2^{\circ} \times 3^{\circ} \times 34$ hybrid vertical levels $\times 3$h.
\end{itemize}
Note, while \emph{OrigRes} is close to the original data resolution, it is not exact -- we shift the time steps in comparison to CarbonTracker by $1.5$h (except for fluxes) and we still regrid the surface fluxes, which had been optimized at $1^{\circ} \times 1^{\circ}$ in CarbonTracker. In addition, in CarbonTracker North America, the full atmosphere is modeled at this higher resolution over a zoomed window in North America. We deliberately chose the horizontal resolution such that LowRes (MidRes) horizontal fields have $32\times 64$ ($64\times 128$) pixels, which is ideal for most modern deep neural network architectures from computer vision \cite{rasp.etal_2024}.



\subsubsection{Data preprocessing}
In order to prepare the three deep learning-ready dataset versions, we introduce a preprocessing chain. Through this chain, we aim to standardize dataset format and ensure that the processed data is directly useable to implement offline tracer transport emulators in the spirit of eq.~\ref{eq:emulation}. Furthermore, the chain enables future work to leverage the presented neural networks on datasets from other transport models. We perform the following preprocessing steps:

\begin{enumerate}
    \item Horizontal regridding: intensive meteorological variables with bilinear interpolation, extensive quantities (\ce{CO2} mixing ratio and air mass) are divided by cell area, and then, alongside \ce{CO2} surface fluxes regridded with conservative interpolation.
    \item Conversion to standard units and variables: masses in $\lbrack Pg \rbrack$, Cconcentrations as ppm mass mixing ratio $\lbrack\frac{10^{-6}kg\ce{CO2}}{kg\ce{DryAir}}\rbrack$, fluxes as $\lbrack\frac{kg\ce{CO2}}{m^2s}\rbrack$, pressure in $\lbrack hPa \rbrack$. We aggregate surface fluxes into ocean, land and anthropogenic fluxes, where the former two would be optimized during an inversion and the latter one prescribed.
    \item Vertical aggregation: pressure weighted mean for intensive quantities, sum for extensive quantities (masses).
    \item Temporal resampling: linear resampling to target resolution.
    \item Flux staggering: surface fluxes are staggered, such that they represent the mean flux between a time step and the next time step.
    \item Flux mass correction: anthropogenic surface fluxes are corrected, such that any mass conservation errors introduced through preprocessing are removed and the mass difference between two time steps matches exactly the surface fluxes.
    \item Temporally splitting into independent training, validation and testing datasets.
    \item Deep learning-optimized storage: we store our dataset in Zarr files, with chunking that optimizes loading of all data at a single time step: We store two arrays per time step, one with all 2D fields and one with all 3D fields.
    \item Statistics: we compute mean and std. dev. statistics for all fields and for all per-level temporal deltas of all fields.
\end{enumerate}

The preprocessing routines are implemented as part of the Neural Transport Python library (\url{https://github.com/vitusbenson/neural_transport}).

\subsubsection{ObsPack station data}
The NOAA ObsPack GLOBALVIEWplus product \cite{schuldt.etal_2023} collects measurements of atmospheric \ce{CO2} from many different scientific laboratories around the globe with instruments at ground-based stations and towers and onboard ships, aircraft and weather balloons. In this study, we use all measurements flagged as representative from the \emph{v9.1\_2023-12-08} product. We compare these \ce{CO2} measurements with our modeled data by extracting the grid cells closest to the horizontal (lat/lon) and vertical position (geopotential height) of the measurement and averaging over 6h time windows. We use the exact same method to extract station time series from the target CarbonTracker data, as we use for the AI models. This allows for an absolute comparison point: the target CarbonTracker data does not achieve perfect prediction of the ObsPack data, meaning we can compare the performance of AI models directly with TM5, the transport model used in CarbonTracker. In future work, the ObsPack station data does also allow for cross-dataset comparison. Note, however, if AI models trained on two different datasets are compared, differences in performance may also stem from the differences in the prescribed surface fluxes, meteorology and initial conditions, and not merely from the learned transport model.

\subsubsection{Evaluation}
We evaluate models by performing quarterly forward runs starting on Jan 1st, April 1st, etc. and running for 3 months each. We then average statistics over the full test period (2018--2020) and compute a range of performance metrics, such as RMSE, $R^2$, decorrelation time (\#days with $R^2 > 0.9$), RMS mass error, relative mean and relative variability. We compute these metrics over individual spatial and temporal coordinate axes and also over sets of axes, to obtain a full picture.

\subsection{Neural Networks}

In this section we describe the neural networks studied in this work. We restrict ourselves to a rather conceptual description and refer the reader to the original papers for in-depth explanations of each architecture. In addition we report the adjustment to the original architectures which we introduce in this work to enable their applicability to atmospheric transport modeling. 

\subsubsection{Motivation}

Atmospheric transport modeling requires processing high dimensional data: at the coarsest resolution, our model input has $32 \times 64 \times (10 \times 10 + 7) \approx 220k$ dimensions (and $\sim 20k$ output dimensions). At such scales, training a standard 2-layer neural network, the multi-layer perceptron (MLP), becomes computationally intractible. In deep learning this challenge is typically approached by introducing inductive biases, that allow to significantly reduce the dimensionality of each matrix multiplication. In this study, from the vast variety of available architectures, we pick four that are representative of generic architectural classes and that previous work has found successful at emulating weather and climate data.

\begin{figure}
    \centering
    \includegraphics[width = \textwidth]{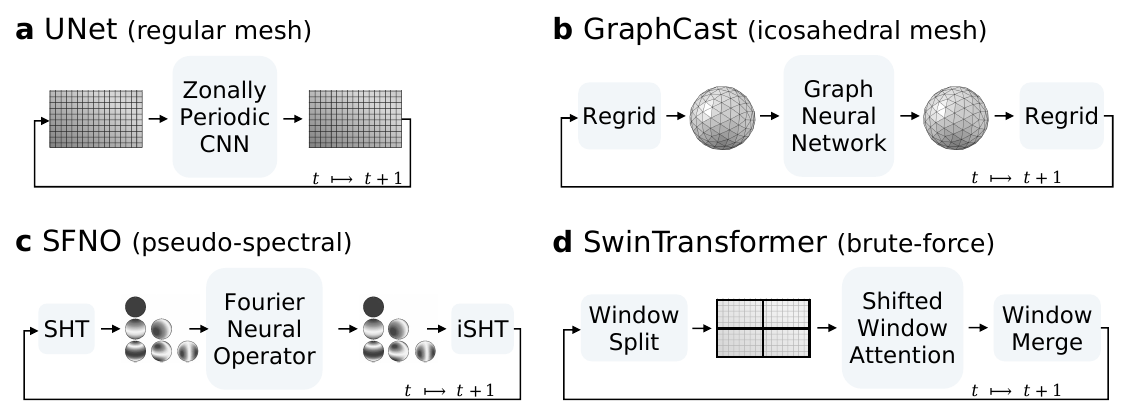}
    \caption{Conceptual depiction of the four deep neural networks included in this study.}
    \label{fig:models}
\end{figure}

Moreover, three out of the four networks coincide with general classes of conventional numerical methods (compare fig.~\ref{fig:models}): a) UNet uses a regular mesh, like finite difference solvers on regular grids, b) GraphCast uses an icosahedral mesh, again analogous to finite difference solvers, c) SFNO is similar to a pseudo-spectral solver, only d) SwinTransformer is unconventional in the way that it favors a brute-force split-process-combine approach, with little resemblance to conventional numerical methods, i.e. it has the least inductive bias.

\subsubsection{Vertical discretization}

In all four approaches, we only consider inductive biases for the horizontal dimension, in the vertical direction we stack all data along the channel dimension and feed that as input. In other words, the models receive an array of values (for forcing, tracers and surface fluxes) per horizontal grid cell, and then process these in a latent space, allowing for vertical mixing and interactions across variables. This approach is independent of the partical vertical discretization pertinent in the data.

In this work, we use CarbonTracker data, which comes at hybrid model levels. Hybrid levels interpolate smoothly between a terrain-following component in the lower troposphere (close to the surface) and constant pressure levels in the upper stratosphere. More specifically, the pressure of each vertical layer is an affine transformation of the surface pressure (which varies with orography).

\subsubsection{UNet}
UNets \cite{ronneberger.etal_2015} are fully convolutional neural networks (CNNs) consisting of an encoder and a decoder arranged in a \emph{U-shape} -- referring to gradual spatial downsampling and subsequent upsampling. We employ UNets that treat the globe as a cylinder, having periodic convolutions in zonal (longitude) direction and zero-padded convolutions at the poles \cite{scher_2018, rasp.etal_2020}. Vertical layers and different variables are simply stacked along the channel dimension.

Our UNet has 4 stages within the encoder and decoder, each consiting of two $3\times 3$ 2d conv layers, that are followed by LeakyReLU and BatchNorm layers and a residual connection. Spatial downsampling is achieved through $2\times 2$ MaxPooling and upsampling through $2\times2$ nearest interpolation. In the first encoder stage, we use a single $7\times 7$ conv layer instead. We add skip connections between the encoder and decoder stages. The network operates on input sizes that are divisible by $16$, but through bilinear upsampling in the first and nearest downsampling in the last layer, we allow for other input shapes as well.

\subsubsection{SwinTransformer}
SwinTransformers \cite{liu.etal_2021b} are transformer neural networks processing 2D inputs by attention between embeddings of windows, which are shifted in each layer. We allow for periodic shifts in zonal (longitude) direction, retain processing at the highest resolution (no hierarchical layers) and adopt relative positional encoding, three architectural design choices which have been proven useful for weather forecasting \cite{willard.etal_2024}.

Our SwinTransformer has 12 layers each consisting of a Multi-head Self-Attention block followed by LayerNorm and a pixelwise MLP (with GELU activation and LayerNorm) and residual connections between blocks. The self-attention is masked in such a way, that only attention within windows of nearby pixels is computed, we use $4 \times 8$ pixel windows. Windows are shifted by half their size at every second layer, with zonally periodic shifts. In contrast to previous work we found using patch embedding to introduce artifacts at longer rollouts, which is why our model directly operates at pixel level (i.e. in $1\times1$ patches). Input shapes need to be divisible by the window shape, we allow for other input shapes through nearest interpolation.

\subsubsection{GraphCast}
GraphCast is a graph neural network (GNN) tailored for weather forecasting. It follows an encode-process-decode layout \cite{battaglia.etal_2018}, with the encoder and decoder mapping between the regular grid (lat-lon) and an icosahedral mesh \cite{keisler_2022}. Thus, they are responsible for two tasks: first, they perform regridding, akin to conventional regridding tools, but here learned, and second, they map the input data into an high-dimensional latent space, as typical for deep neural networks. On the icosahedral mesh in latent space, the processor component processes the data to obtain a powerful embedding from which the time delta of the target variables can be extracted. More specifically, the processor uses message passing layers in local neighborhoods of each grid cell with additional long-range connections \cite{lam.etal_2023a}. This can be understood as local stencils on the sphere that process information just like in a conventional finite difference solver, with the addition of some non-local interactions between supernodes, that can further enhance predictions.

Our GraphCast has a processor with 8 layers, each performing message passing between neigboring nodes on an icosahedral multi-mesh that has been refined 3 times (levels 0-3). The encoder uses a bipartite graph to map between the regular grid representation and multi-mesh nodes by assigning all grid cells to a multi-mesh node whose center is less than $0.75$ times the maximum inter-node distance in the level 3 mesh away from that node. The encoder and decoder map between data space and a latent space with 256 channels. Like the original GraphCast we use Swish activations and layer norm. Our message passing layer use a mean operation to aggregate incoming information from neighboring nodes.


\subsubsection{Spherical Fourier Neural Operator}
Spherical Fourier Neural Operators (SFNO) \cite{bonev.etal_2023} are an extension of the Fourier Neural Operator (FNO) \cite{li.etal_2021b} to the sphere, by replacing Fourier transforms with spherical harmonics transforms (SHT). An FNO Block performs channel-wise spatial processing in the spectral domain and combines this with channel-mixing in the grid domain. The SFNO consists of many blocks, each using the SHT and inverse SHT to map between grid and spectral space. We use linear transformations in spectral space and local MLPs in grid space.

\subsection{Details}
We train our deep neural networks using the Neural Transport Python library (\url{https://github.com/vitusbenson/neural_transport}). Our experiment scripts are published in the CarbonBench Python repo (\url{https://github.com/vitusbenson/carbonbench}).

\subsubsection{Optimization}
We train our models with ADAM in a two-stage fashion. First, with a cosine learning rate schedule and linear warm up on next-step prediction. Afterwards with a constant learning rate and a n-steps-ahead schedule, where we iteratively increase the lead time during training every 2 epochs until 31-steps-ahead. For hyperparameter tuning and ablation studies, we do next-step training for 100k steps, and for the final models for 300k steps. In this work, we optimize always against the full 3D \ce{CO2} field from CarbonTracker, future work may consider additionally including a part of the ObsPack measurements (which are only used for evaluation in this work) or weighting targets differently.

\subsubsection{CentFlux}
We scale and shift the model output with the std. dev. and mean of the temporal deltas of each target variable vertical layer. Afterwards, we add the previous time step 3D field to obtain a raw prediction for the next time step. In addition, we add the surface fluxes to the lowest vertical layer. Due to steadily rising anthropogenic emissions, the input \ce{CO2} mean is increasing over time, which would represent a covariate shift, to which neural networks are rarely robust. To account for this, we center the input \ce{CO2} field at each time step to have zero mean. This fix should allow stable transport for arbitrary levels of atmospheric \ce{CO2}. Throughout this manuscript we call the addition of surface fluxes at the lowest vertical level and the centering of \ce{CO2} input fields jointly \emph{CentFlux}.

\subsubsection{SpecLoss}
Previous work identified divergence in the power spectra to be symptomatic for models becoming unstable for longer rollouts \cite{chattopadhyay.hassanzadeh_2023}. To improve in this regard we introduce an additional loss term that regularizes predictions. \emph{SpecLoss} measures the difference in spectral power densities between observed and predicted 2D fields (i.e. at each vertical level). We leverage the spherical harmonics transform to obtain spectral coefficients, from which we compute the spectral power density. Our approach is similar to a regularization term used in NeuralGCM \cite{kochkov.etal_2023}.

\subsubsection{Massfixer}
Tracer transport fulfills the continuity equation, which stems from mass conservation, in other words, the total mass of simulated \ce{CO2} in the atmosphere at $t+1$ should match the mass at $t$ plus the total mass input through the surface fluxes. While some conventional numerical approaches like finite volume methods fulfill tracer mass conservation by design, others, such as semi-lagrangian or pseudo-spectral schemes do not. Also deep neural networks are only softly constrained to fulfill mass conservation (if zero emulation error is achieved, mass is necessarily conserved). Similarly to previous attempts to correct conventional approaches \cite{diamantakis.flemming_2014}, we adopt a simple mass fixer, that scales the predicted mass at each time step by the desired mass calculated from the surface fluxes. This fixer leads to proportionally larger adjustments in grid cells with more tracer mass.

\section{Results}

\begin{figure}
    \centering
    \includegraphics[width = \textwidth]{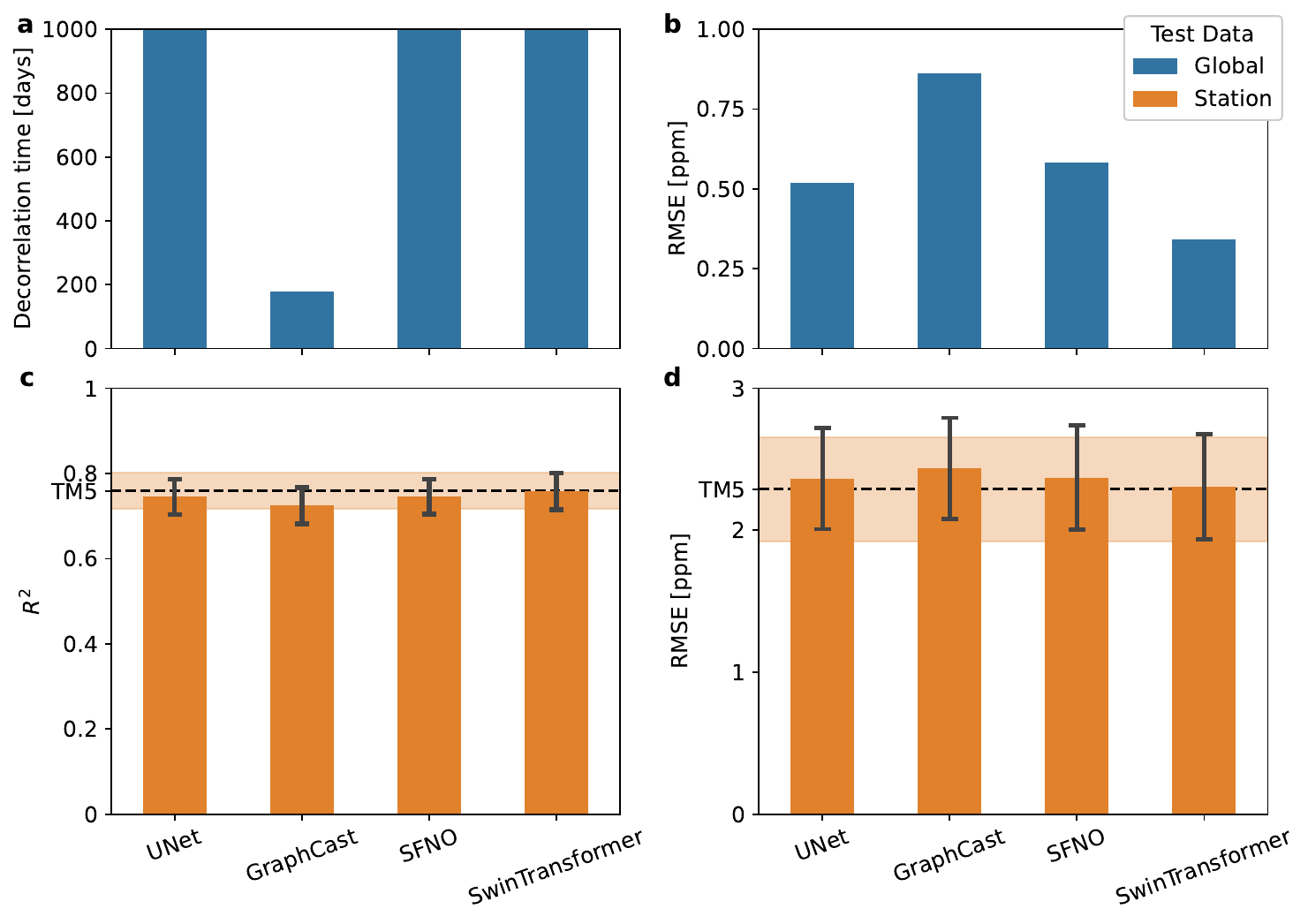}
    \caption{Intercomparison between the best models per architecture. In blue (a\&b), the performance is evaluated by scoring the global predicted 3D field against the ground truth \ce{CO2} field from the test period of the LowRes dataset -- this allows for comparisons between the AI models. In orange (c\&d), the performance is evaluated at ObsPack stations. This allows, in addition, to compare against TM5 (dashed black lines), the transport model used to produce the ground truth dataset. At ObsPack stations, in addition to the mean scores, we also display uncertainty estimates: the std. dev. over stations scaled by the square root of the number of stations. Local $R^2$ (c) and global (b) and local RMSE (d) are computed for quarterly 90-day forward runs, the decorrelation time (a) is estimated from a single 3 year forward run.}
    \label{fig:intercompare}
\end{figure}

\subsection{Model intercomparison}
We evaluate global and local test set performance of the four neural network architectures, each with tuned hyperparameters, and report the results in fig.~\ref{fig:intercompare}. UNet, GraphCast, SFNO and SwinTransformer all achieve stable transport for at least 6 months with local performance almost equal to TM5, that is, to the ground truth that models had been trained on. The best model is SwinTransformer, which achieves a global $R^2$ of $0.99$ over quarterly forecasts, i.e. almost perfect emulation. Performance degrades when looking at the other three models, with UNet $>$ SFNO $>$ GraphCast. Here, GraphCast has more than double the global RMSE compared to SwinTransformer, but still stays below 1 ppm over 90 day forward runs. Furthermore, GraphCast runs become unstable after $178$ days, while the other three models display decorrelation times above 3 years, indicating long-term stability (fig.~\ref{fig:intercompare}a). At station level, the difference are of lower magnitude, but still significant (fig.~\ref{fig:intercompare}c\&d). In the following we assess the performance of the SwinTransformer, the best performing model, in more detail, with the equivalent plots for the other models provided in the supplementary material.

\subsection{Best performing model}

\begin{figure}
    \centering
    \includegraphics[width = \textwidth]{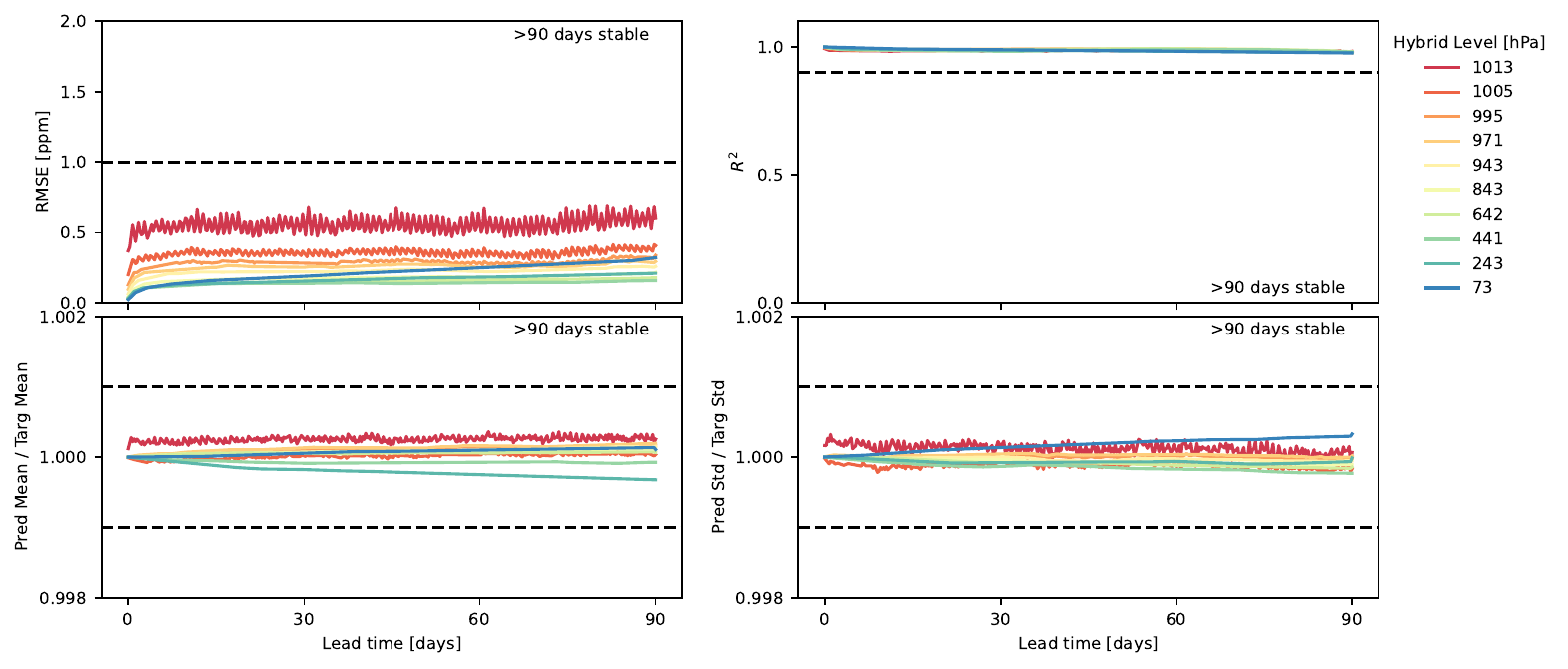}
    \caption{Key metrics per vertical layer for quarterly forecasts over the test set for SwinTransformer. We report metrics per time step and vertical level, i.e. they represent properties of the 2D maps of atmospheric \ce{CO2} mass mixing ratios at different vertical levels. The metrics are averaged over quarterly reset 90-day forward runs. Dashed lines indicate arbitrarily set thresholds which subjectively signify stable simulation (e.g. RMSE $<$ 1 ppm is a goal for many \ce{CO2} MRV systems).}
    \label{fig:keymetrics_swintransformer}
\end{figure}

The SwinTransformer produces stable forecasts in terms of RMSE, $R^2$, relative mean and relative std. dev. over 90 days. Fig.~\ref{fig:keymetrics_swintransformer} compares the performance for different levels. Mostly, the performance varies little for different layers, with the exception that the surface layer has a significantly larger RMSE compared to all other layers (over $1.5$x). Moreover, while in the lower troposphere after a brief annealing phase during the first few forecast steps the predictions are of approximately constant quality, there is a drift with increased performance degradation in the upper stratosphere (the top three layers).

\begin{figure}
    \centering
    \includegraphics[width = \textwidth]{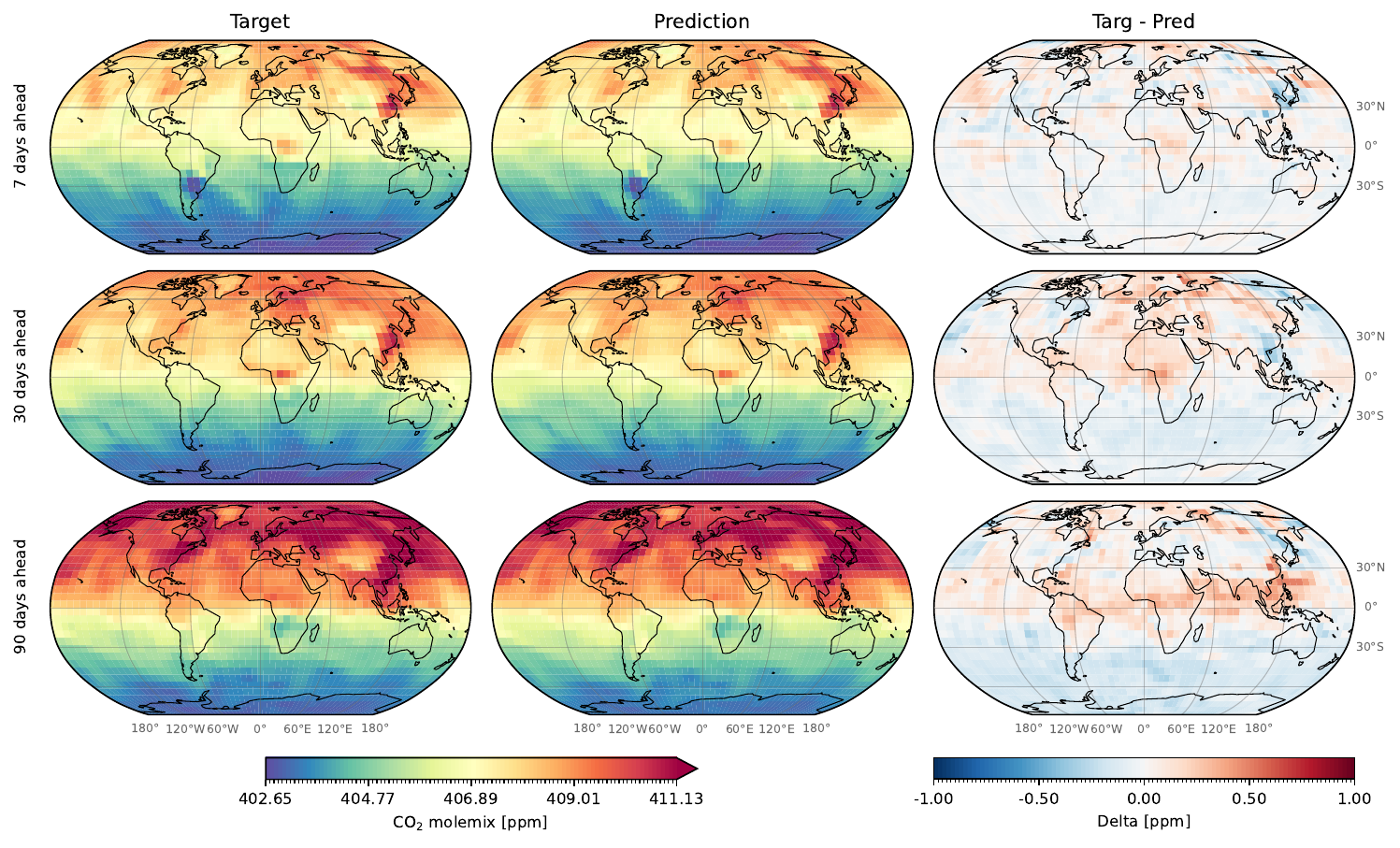}
    \caption{Maps of Total Column \ce{CO2} Target, Prediction by SwinTransformer and Error for different lead times. Shown is a single forward run starting from Jan 1st, 2018. }
    \label{fig:maps_swintransformer}
\end{figure}

Qualitatively, SwinTransformer captures the large-scale motion of \ce{CO2} in the atmosphere, as depicted by maps of total column \ce{CO2} (fig.~\ref{fig:maps_swintransformer}). The largest errors appear in eastern Asia, a region known for large anthropogenic emission. Otherwise, error patterns appear to follow fronts in the atmospheric field, indicating mildly decreased performance over sharper gradients (fig.~\ref{fig:maps_swintransformer}).

\begin{figure}
    \centering
    \includegraphics[width = \textwidth]{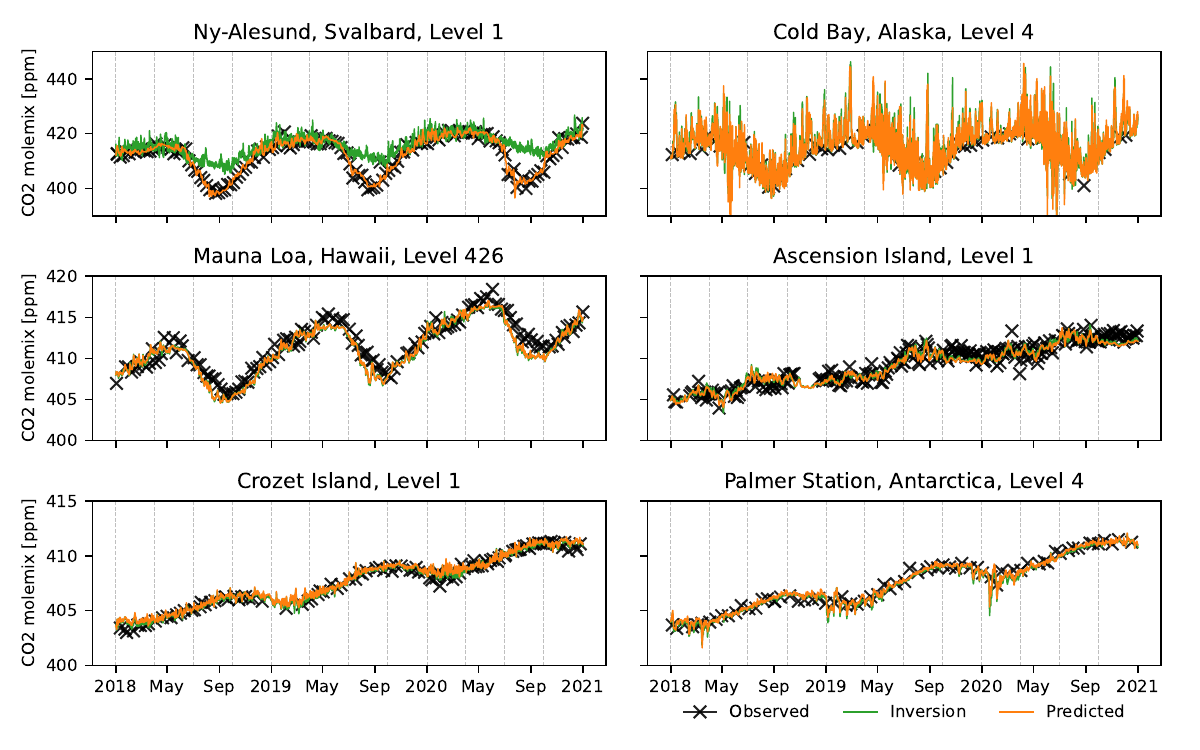}
    \caption{Performance of SwinTransformer (orange line) compared to TM5 (the training target, here: \emph{Inversion}, green line) at six measurement stations from the ObsPack Globalview product. Shown are 90-day forward runs, the light grey lines indicate the dates on which the runs are reset.}
    \label{fig:obspack_swintransformer}
\end{figure}

Zooming in on a few stations from the ObsPack Globalview product, SwinTransformer generally performs similar to the training target TM5 (fig.~\ref{fig:obspack_swintransformer}). Interestingly, for the Svalbard station, SwinTransformer captures the seasonal cycle in the observations well, whereas TM5 oversmoothes it. There are barely any jumps visible at the quarterly intervals (grey dotted lines), where the SwinTransformer initial state is reset. This is in line with the previous result, that SwinTransformer displays little performance degradation over 90 day horizons. While it is unclear exactly why SwinTransformer outperfroms TM5 in Svalbard, it may be related to the stations vicinity to the poles and differences in the boundary layer vertical transport of the two models.

\subsection{Mass Conservation}

\begin{figure}
    \centering
    \includegraphics[width = \textwidth]{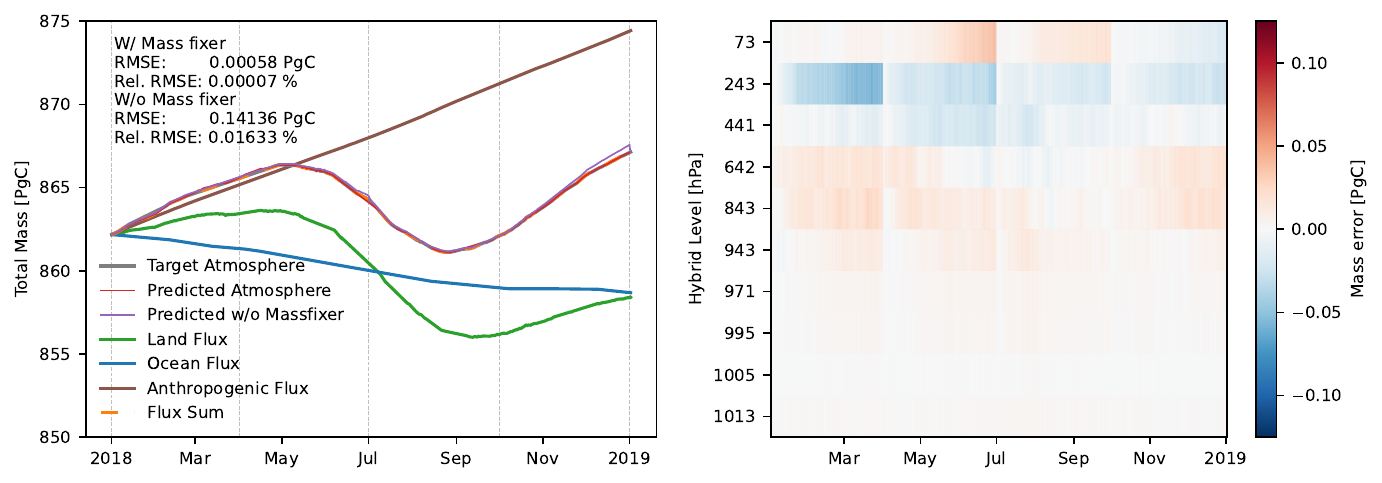}
    \caption{Mass Conservation of SwinTransformer globally (left) and per level (right). In the left panel, the total mass in the target atmosphere (grey line) and in the predicted atmosphere (red line) match exactly with a cumulative sum of the surface fluxes (orange dotted line), i.e. they are plotted on top of each other indicating mass conservation. The flux sum is the sum of the Anthropogenic (brown), Land (green) and Ocean (blue) fluxes. In addition, we show performance without the massfixer (purple line). The right panel shows the difference of the total mass per level and time step between the SwinTransformer prediction (after applying the massfixer) and the target.}
    \label{fig:mass_swintransformer}
\end{figure}

Fig.~\ref{fig:mass_swintransformer} presents global and per-level mass conservation results with SwinTransformer. Globally SwinTransformer with the mass fixer achieves an RMSE of $0.00058$ PgC, which may be considered neglible in comparison to the total atmospheric mass of $\sim865$ PgC in 2018. This remaining mass error likely stems from numerical problems: our deep neural networks operate with 32-bit floating points, which can give performance issues especially when dealing with division of relatively large numbers. Notably, the mass fixer greatly enhances the conservative properties of SwinTransformer in comparison to the free-running neural network (purple line, fig.~\ref{fig:mass_swintransformer} left side): it has over $0.01$\% relative mass RMSE, which particularly manifests in an overprediction of mass in november and december.

Analyzing the mass error per vertical layer gives insight into the vertical transport learned by SwinTransformer. Fig.~\ref{fig:mass_swintransformer}, right side, indicates that the upward vertical transport is too weak in northern hemisphere winter (too little mass in upper stratosphere) and too strong in summer. Notably, vertical transport in the lower layers close to the surface displays little mass error, albeit those layers being more heavily influenced by diurnal variability and surface fluxes.

\subsection{Long-term Stability}

\begin{figure}
    \centering
    \includegraphics[width = \textwidth]{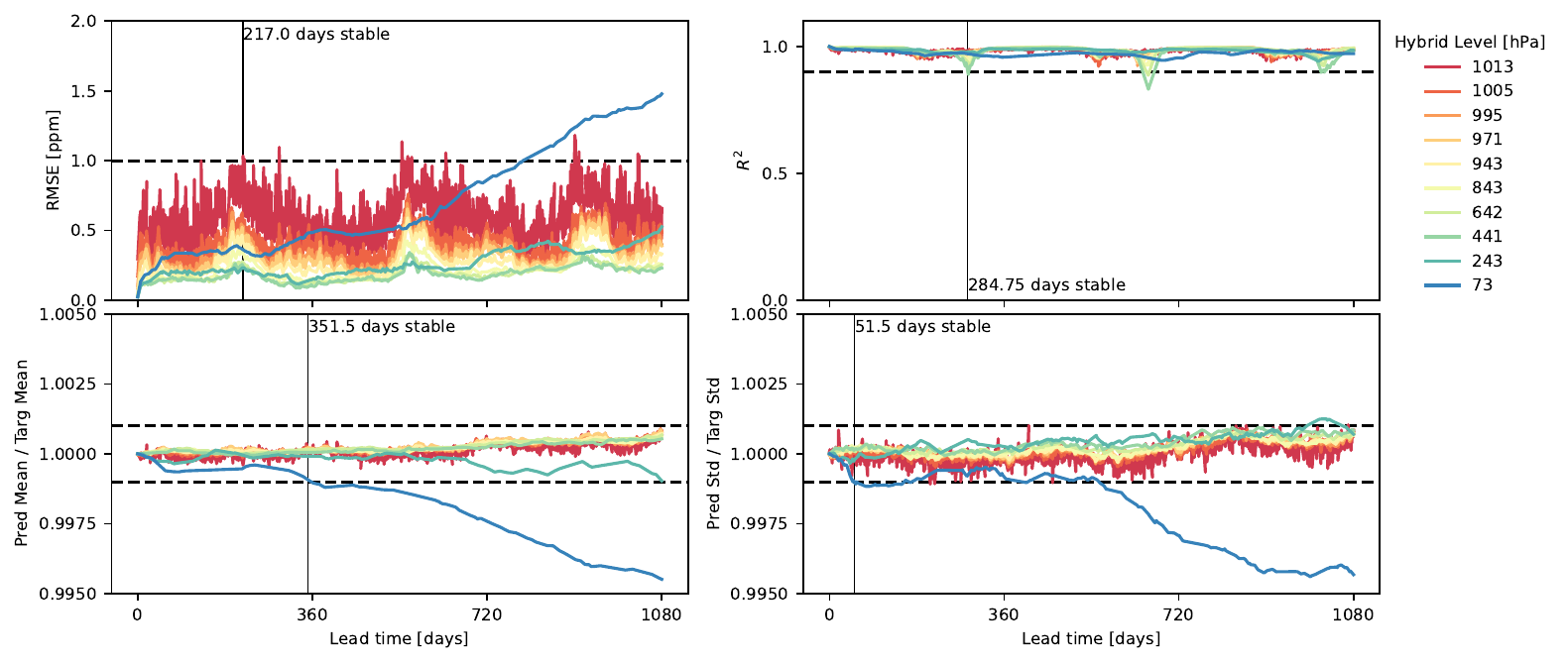}
    \caption{Key metrics per vertical layer for a single 3-year rollout with SwinTransformer starting from Jan 1st, 2018. As in fig.~\ref{fig:keymetrics_swintransformer}, we report metrics per time step and vertical level. The metrics are averaged over quarterly reset 90-day forward runs. Dashed lines indicate arbitrarily set thresholds.}
    \label{fig:longmetric_swintransformer}
\end{figure}

\begin{figure}
    \centering
    \includegraphics[width = \textwidth]{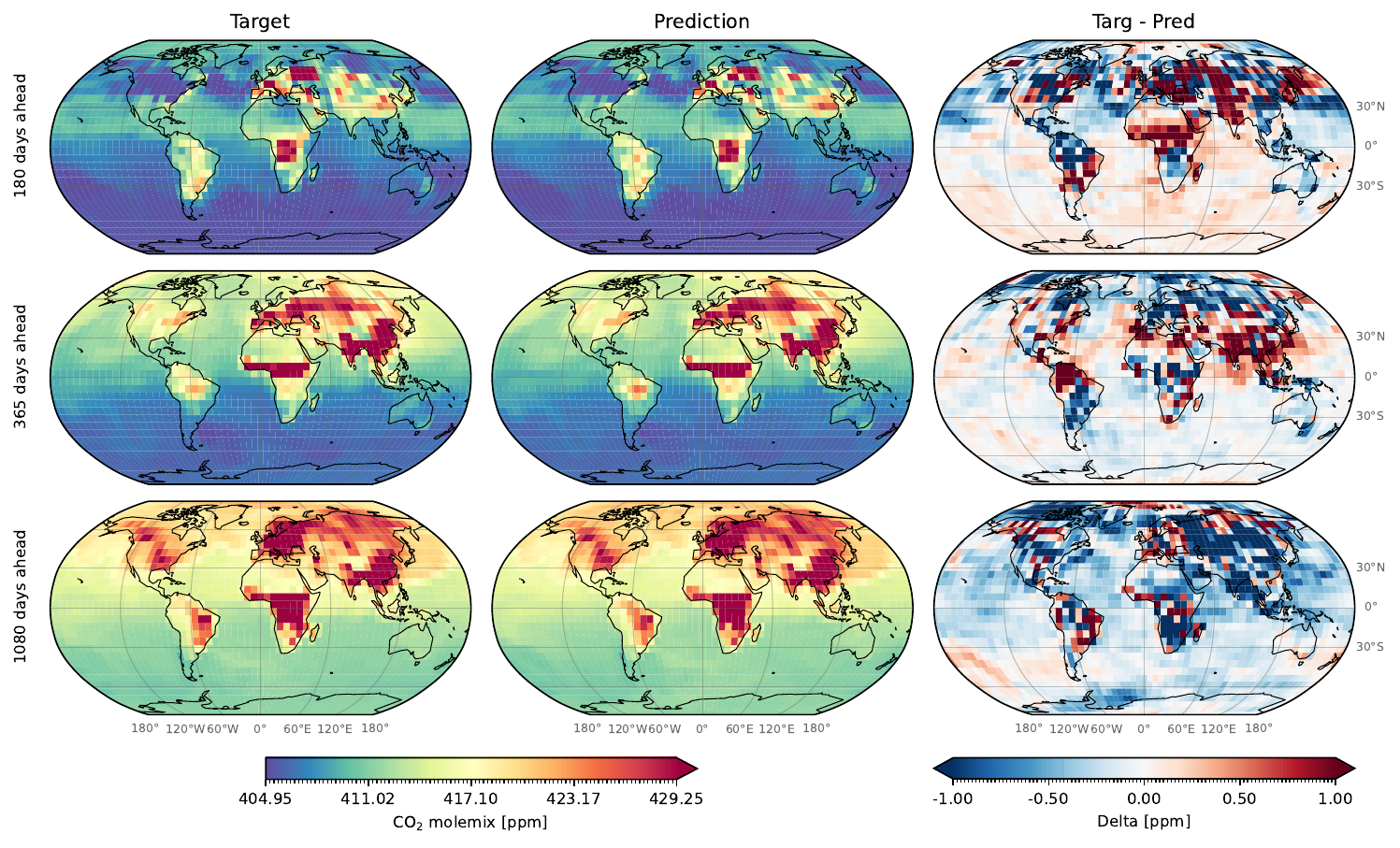}
    \caption{Maps of surface layer ($1013$ hPa in a standard atmosphere) \ce{CO2} Target, Prediction by SwinTransformer and Error for different lead times of a 3-year rollout starting from Jan 1st, 2018.}
    \label{fig:longmap_swintransformer}
\end{figure}

While this paper mostly focuses on prediction horizons up to $90$ days, we also performed a 3-year rollout of the SwinTransformer over the full test period. SwinTransformer remains stable even after over 3 years rollout, but starts to display errors above $1$ppm in many regions (fig.~\ref{fig:longmap_swintransformer}).

More specifically, the surface layer RMSE first crosses $1$ ppm after $217$ days (fig.~\ref{fig:longmetric_swintransformer}) and the RMSE near the surface generally displays cyclical behavior, with highest errors in northern hemisphere summer. The highest layer, representing the upper stratosphere, is unstable over rollout time: it is being oversmoothed and accumulates too little mass over time. For most inverse modeling purposes, this is of lesser concern, as the upper stratosphere contains less carbon and there are typically no direct measurements of \ce{CO2} taken at such altitude.

Overall the results are particularly promising as previous work has repeatedly noted challenges in the stability of long-term rollouts of neural network-based PDE emulators \cite{brandstetter.etal_2022a, bonev.etal_2023, lippe.etal_2023a}. Moreover, \ce{CO2} transport may be considered particularly challenging as atmospheric \ce{CO2} concentrations keep rising, naturally pushing the distribution of the atmospheric tracer field away from the training distribution and constituting an out-of-domain (OOD) problem. Still, future work needs to assess the robustness of our models to distribution shifts beyond the rise in \ce{CO2} during the test set. For example considering generalization to significantly different surface fluxes could be relevant. While preliminary experiments with transporting zeroed-out surface fluxes indicated no non-physical behavior, caution needs to preside and thus extrapolation far from training data may be a limitation of the transport emulator. 

\subsection{Differences between AI model architectures}

\begin{figure}
    \centering
    \includegraphics[width = \textwidth]{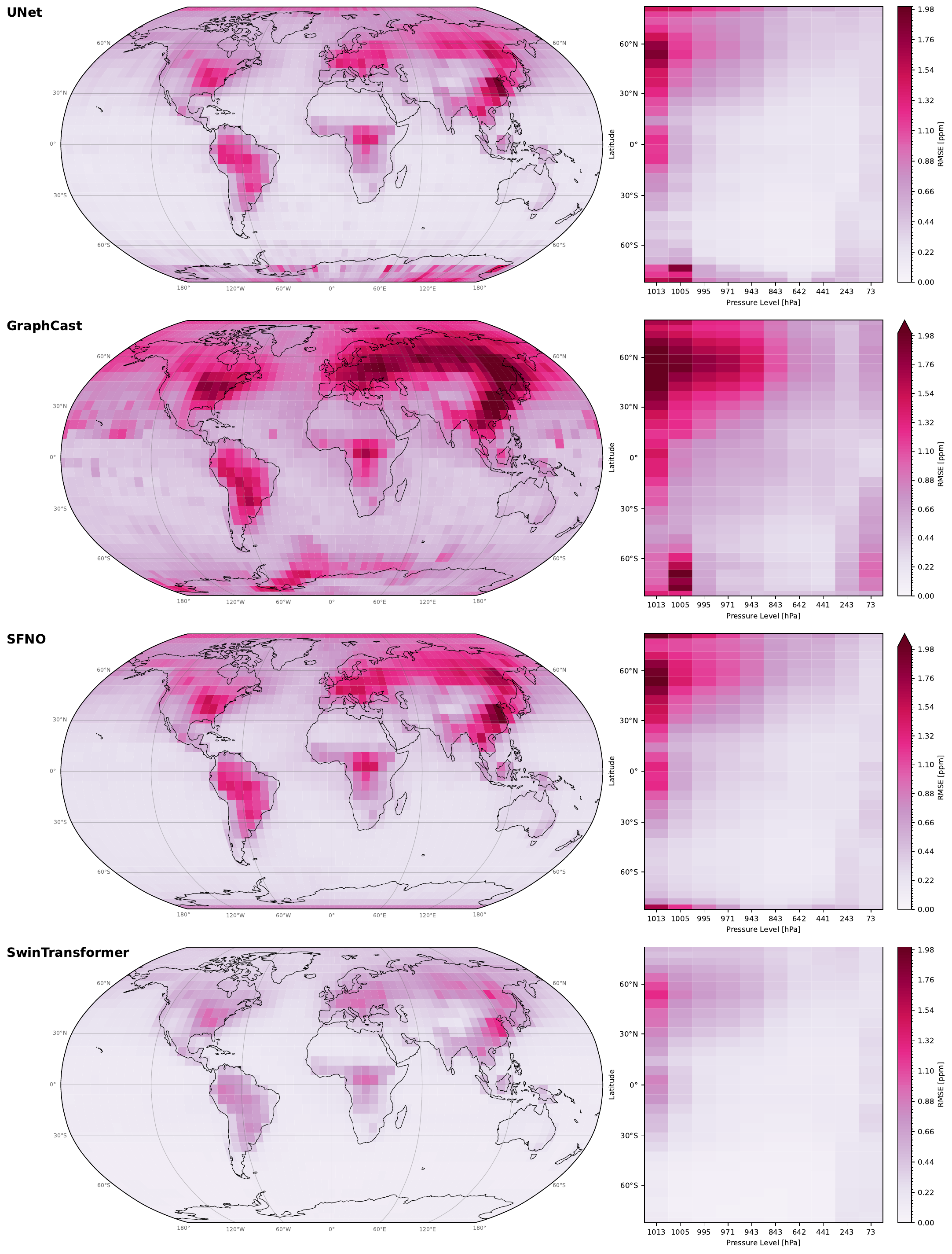}
    \caption{RMSE patterns of the four AI models. For each model, shown is the RMSE per horizontal grid cell averaged over time and vertical level (left side) and per latitude and vertical level averaged over time and longitude (right side). Scores are for quarterly 90-day forward runs.}
    \label{fig:rmsepatterns}
\end{figure}

The four AI models included in this study build on different underlying principles (mesh-based vs. pseudo-spectral vs. brute-force). Hence it is less surprising that there are differences in the patterns of model residuals between models. Fig.~\ref{fig:rmsepatterns} presents RMSE patterns. For all models, RMSE seemingly scales with \ce{CO2} variability: regions with large biosphere dynamics such as the tropics or boreal forests, and areas with large anthropogenic emissions such as eastern Asia in the near-surface layers have consistently larger errors. UNet, SFNO and GraphCast all have higher errors at the poles. For SFNO it is very limited to the pole grid cell itself, likely because the spherical harmonics there do not allow for zonal variability. For UNet, the impact is a bit larger, mirroring the smoothing effect of convolutions with zero padding at the poles. GraphCast has the most severe problems with the poles. This might be related to the encoder and decoder of GraphCast, which map between grid cells on the regular grid and nodes on the icosahedral mesh. Near the poles, many grid cells are mapped to a single node, which could potentially result in stability problems.

\begin{table}[t]
    \centering
    \begin{tabularx}{\textwidth}{Xcccccc}
    \toprule
    Model & CentFlux & SpecLoss & \#Params & Decorr Time & $R^2$ & RMSE \\
    \midrule
    UNet S & \xmark & \xmark & 9.6M& $1.5$ & $0.07$ & $>100$ \\
    UNet S & \cmark & \xmark & 9.6M& $>90$ & $0.98$ & $0.57$ \\
    \underline{UNet S} & \cmark & \cmark & 9.6M& $>90$& $0.98$ & $0.52$ \\
    UNet XS & \cmark & \cmark & 2.7M & $>90$ & $0.98$ & $0.62$ \\
    UNet M & \cmark & \cmark & 35.7M & $>90$ & $0.98$ & $0.52$ \\
    \midrule
    GraphCast XS & \xmark & \xmark & 5.2M & $41.25$ & $0.87$ & $1.63$ \\
    GraphCast XS & \cmark & \xmark & 5.2M & $>90$ &  $0.95$& $0.96$ \\
    \underline{GraphCast XS} & \cmark & \cmark & 5.2M & $>90$ & $0.96$ & $0.86$ \\
    GraphCast XXS & \cmark & \cmark & 1.3M & $>90$ & $0.95$ & $0.92$ \\
    GraphCast S & \cmark & \cmark & 8.8M & $>90$ & $0.96$ & $0.87$ \\
    GraphCast XS mesh=0--2 & \cmark & \cmark & 5.2M & $>90$ & $0.94$ & $0.99$ \\
    \midrule
    SFNO M & \xmark & \xmark & 35.7M & $>90$ & $0.97$ & $0.67$ \\
    SFNO M & \cmark & \xmark & 35.7M & $>90$ & $0.98$ & $0.59$ \\
    \underline{SFNO M} & \cmark & \cmark & 35.7M & $>90$ & $0.98$ & $0.58$ \\
    SFNO S & \cmark & \cmark & 8.9M & $>90$ & $0.98$ & $0.59$ \\
    SFNO L & \cmark & \cmark & 53.5M & $>90$ & $0.98$ & $0.59$ \\
    \midrule
    SwinTransformer M & \xmark & \xmark & 37.9M & $>90$ & $0.97$ & $0.79$ \\
    SwinTransformer M & \cmark & \xmark & 37.9M & $>90$ & $0.99$ & $0.37$ \\
    \textbf{\underline{SwinTransformer M}} & \cmark & \cmark & 37.9M & $>90$ &  $0.99$ & $0.34$ \\
    SwinTransformer S & \cmark & \cmark & 6.4M & $>90$ & $0.99$ & $0.36$ \\
    SwinTransformer L & \cmark & \cmark & 85.2M & $>90$ & $0.99$ & $0.34$ \\
    SwinTransformer M ps=4 & \cmark & \cmark & 38.8M & $>90$ &  $0.97$ & $0.70$ \\
    \bottomrule
    \end{tabularx}
    \caption{Ablation study highlighting the best configuration per model architecture (underline) and the best overall model (bold). For each model, we compare three different sizes, whether to center the input \ce{CO2} field to account for covariate shift and to add surface fluxes directly to the lowest vertical layer (Centering \& Flux Addition, i.e. \emph{CentFlux}), and, whether to leverage an additional loss term which measures divergence in the spectral power densities (\emph{SpecLoss}). For GraphCast, we additionally ablate the resolution of the icosahedral multi-mesh (\emph{mesh}, default is 0--3), and for SwinTransformer, we ablate the patch size (\emph{ps}, default is 1). We report three metrics: decorrelation time, $R^2$ and RMSE - all over 90-day forward runs.}
    \label{tab:ablation}
\end{table}

\subsection{Ablations}
In our experiments we found the four AI models to not work very well for \ce{CO2} prediction out-of-the-box. Especially the mesh-based methods UNet and GraphCast displayed issues with low stability over longer rollouts. In contrast, the final models presented in this paper are stable and mass-conserving for over $90$ days. Table~\ref{tab:ablation} presents insights into the design choices that lead to the improved performance on the LowRes dataset.

For each of the four models, we ablate the model size and two training tricks that particularly increased the stability. The first one, \emph{CentFlux}, is a combination of centering the 3D \ce{CO2} fields and adding the prescribed surface fluxes to the lowest vertical layer. The second one, \emph{SpecLoss}, is an additional loss term that penalizes deviations in the spectral power spectrum (computed with the spherical harmonic transform) between the model output and the target.

For all models, \emph{CentFlux} is essential to achieve stable rollouts and improves the performance significantly. \emph{SpecLoss} additionally enhances scores, but the gains are smaller. The four models have different optimal model sizes. While the best GraphCast in our experiment (size XS), has 5.2M parameters, the best UNet (size S) has 9.6M, and the best SFNO and SwinTransformer (both size M) have 35.7M and 37.9M parameters respectively. Note, models with more parameters do not necessarily have better performance: UNet outperforms SFNO slightly on RMSE. Still, that in our experiments it was significantly more challenging to scale GraphCast to larger size compared to SwinTransformer is probably one of the reasons why it is the worst model architecture in our intercomparison.

\subsection{Computational Costs}

\begin{figure}
    \centering
    \includegraphics[width = 0.6\textwidth]{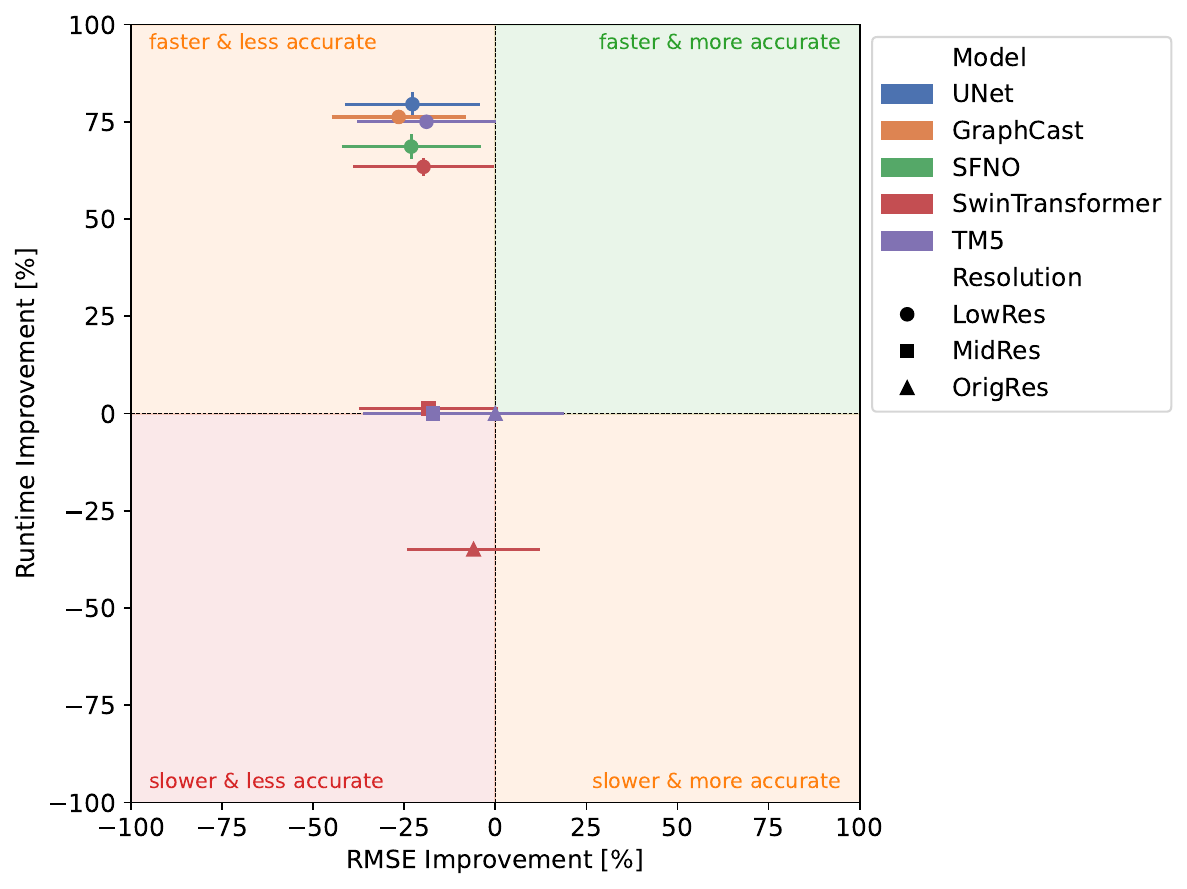}
    \caption{Pareto frontier of inference runtime vs. model performance (RMSE at ObsPack Globalviewplus stations), plotted relative to the TM5 OrigRes target data. Runtime improvements are excluding IO and based on estimated TM5 runtimes. }
    \label{fig:speed}
\end{figure}

One reason why AI-based emulators of ERA5 have garnered interest is because they offer significant speed-ups over conventional NWP models at inference time. Conceptually these speed ups arise as most ERA5 emulators use $10$x less vertical layers (13 instead of 137), $30$x higher time step ($6$h instead of $12$min), purely explicit solvers (no iterative scheme for implicit steps necessary) and compute accelerators (GPUs/TPUs instead of CPUs). These speed ups at inference time come with a trade-off: first significant compute resources need to be allocated in order to train the models. For NWP such an investment is often quickly justified, as many model runs are necessary every day.

Naturally, one may wonder if a parallel to neural network emulation of atmospheric transport can be drawn. However, the SwinTransformer is not significantly faster compared to TM5 (fig.~\ref{fig:speed}). Performance here is highly hardware dependent, but as a rough estimate SwinTransformer takes $\sim1.5$sec for a 30 day forward run on a single Nvidia A40 GPU. Fig.~\ref{fig:speed} compares the speed of the four AI models and TM5 at different resolutions. For this, we measured the model time in an idealized scenario, removing all pre- and postprocessing of model inputs and outputs, and instead directly reading and writing the raw tensors from memory. We then measure the speed of 30 day forecasts with 10 repetitions on a Nvidia A40 GPU. Generally, we notice only small differences between the AI models.

Running the TM5-MP model, which improves upon TM5 through OpenMPI parallelization \cite{williams.etal_2017}, takes $\sim8$ minutes on a machine with 24 CPUs for a 1 month forward run on a $3^{\circ} \times 2^{\circ}$ grid and $\sim2$ minutes on $6^{\circ} \times 4^{\circ}$ \cite{segers.etal_2020}. We assume 50\% time is spent in IO and plot estimated runtimes for TM5 without IO in Fig.~\ref{fig:speed}, with OrigRes and MidRes runs to take $4$ minutes and LowRes to take $1$ minute on a single modern machine with 24 CPUs.

The lack of speed-up can possibly be explained with a number of factors. First, TM5 is run on a $2^{\circ}\times3^{\circ}$ grid, which does not require an extremely small time step. Second, TM5 uses about the same number of vertical layers as SwinTransformer. Third, tracer transport in TM5 is entirely linear (in the surface fluxes), and the mass fluxes for each grid cell are pre-computed. After this is done, transport becomes cheap. Fourth, while TM5 still does not run on GPUs, it reaps a number of benefits from its maturity, such as leveraging fast FORTRAN code and parallelization through OpenMPI.

\section{Discussion}
In this work, we trained deep neural networks to emulate the atmospheric transport of \ce{CO2}. We test four models and find SwinTransformer to perform best, with almost perfect emulation for 90 days, and stable and mass-conserving emulation for multiple years ahead. For this we adjust the model architecture, decoupling the drift in \ce{CO2} from its dynamics by leveraging centered \ce{CO2} fields as inputs and using a post-hoc flux scheme to correct the mass balance. Yet, the presented model is not giving large computational advantages compared to conventional approaches, at least not at low resolution. 

Storm-resolving models allow for explicit treatment of convection, with large impact on vertical transport of air masses and \ce{CO2}. Some modeling centers are already experimenting with storm-resolving transport model runs \cite{agusti-panareda.etal_2014, gelaro.etal_2015, agusti-panareda.etal_2022a, agusti-panareda.etal_2023}, which typically require to run an online transport model. Here, AI models could leverage model output and offer an alternative route ahead.

Considering higher resolution might offer room for speed ups: doubling the horizontal resolution of conventional solvers increases the computation costs by roughly $10$x \cite{hoefler.etal_2023}. Yet, some of the errors of transport representation in current inverse modeling schemes are attributed to low resolution \cite{remaud.etal_2018, agusti-panareda.etal_2019}. Hence, developing multi-resolution training schemes, e.g. by utilizing the cross attention mechanism\cite{jaegle.etal_2021, jaegle.etal_2022, alkin.etal_2024, serrano.etal_2024a}, which is straight-forward with the data in CarbonBench, may enable more accurate low-resolution models that are still computationally feasible for inverse modeling by emulating the high resolution solvers. Moreover, modeling the atmosphere in a highly compressed space may yield further improvements \cite{han.etal_2024}, for instance, such a transport model could render the usage of full resolution wind fields from ERA5 feasible.

Furthermore, there is still a lot of room for common techniques used to speed up AI models. Model distillation is a technique to significantly reduce the parameter count of neural networks without loosing much in terms of skill. Quantization leverages lower numerical precision to decrease memory footprint and increase speed. On a programming language level, just-in-time compilation, e.g. through torchscript, can speed up certain operations. And more generally, data loading can be optimized through asynchronous techniques, clever caching and parallelization.

Future work may also explore the applicability of the neural network solvers for inverse modeling, that is inferring surface fluxes from observed atmospheric measurements. The implementation of the neural networks is fully differentiable, which opens new avenues for obtaining the sensitivities required for the inversions. Furthermore, some inverse modeling approaches rely on the creation of large ensembles. Since neural networks natively support batched processing, there is potential for speed ups (generating a full ensemble can be as cheap as a single forward run).


%
%

\section*{Open Research Section}
We construct the CarbonBench dataset from existing open data from CarbonTracker North America version CT2022 \cite{jacobson.etal_2023a} (\url{http://doi.org/10.25925/z1gj-3254}) and from ObsPack GLOBALVIEWplus \ce{CO2} v9.1 \cite{schuldt.etal_2023} (\url{http://doi.org/10.25925/20231201}). We provide code that downloads the data efficiently from the original data providers and processes it into the formats used in this study, yet in line with the original data licenses we do not re-distribute the data.

We publish the code to run all our experiments and reproduce the results in this paper in the CarbonBench Python repo (\url{https://github.com/vitusbenson/carbonbench}). The deep neural networks are implemented in the Neural Transport Python library (\url{https://github.com/vitusbenson/neural_transport}), a versatile software package containing dataset creation, data loading, training and evaluation routines intended to be easily usable in other research projects.

\acknowledgments
VB is grateful for stimulating discussions to Fabian Gans, Maximilian Gelbrecht, Martin Heimann, Martin Jung, Albrecht Schall, Sam Upton and many others at MPI Jena and ETH Zürich. AW, CR \& MR acknowledge funding by the European Research Council (ERC) Synergy Grant Understanding and modeling the Earth System with Machine Learning (USMILE) under the Horizon 2020 research and innovation programme (Grant agreement No. 855187).

\clearpage

\ifAGU
\else
\section{Supplementary Information}


\begin{figure}[!ht]
    \centering
    \includegraphics[width = \textwidth]{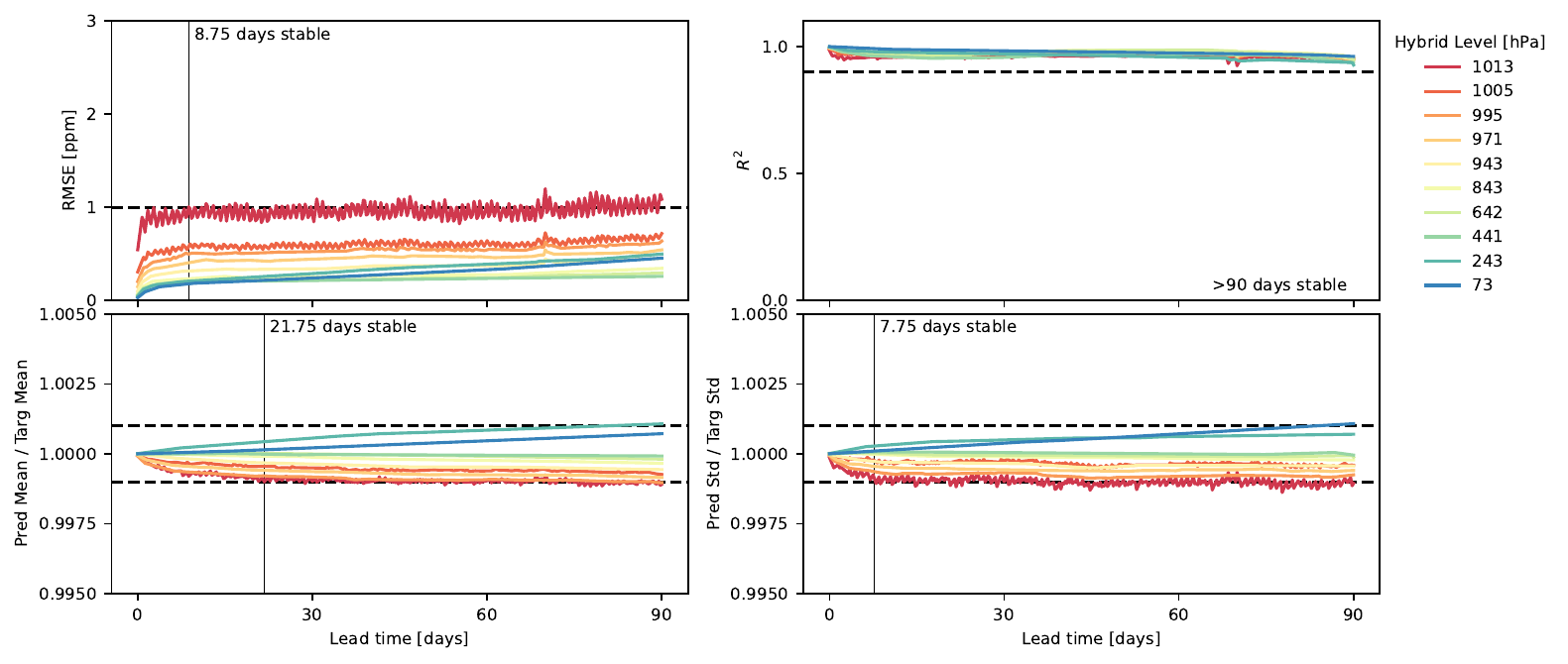}
    \caption{Same as fig.~\ref{fig:keymetrics_swintransformer} but for UNet.}
    \label{fig:keymetrics_unet}
\end{figure}

\begin{figure}
    \centering
    \includegraphics[width = \textwidth]{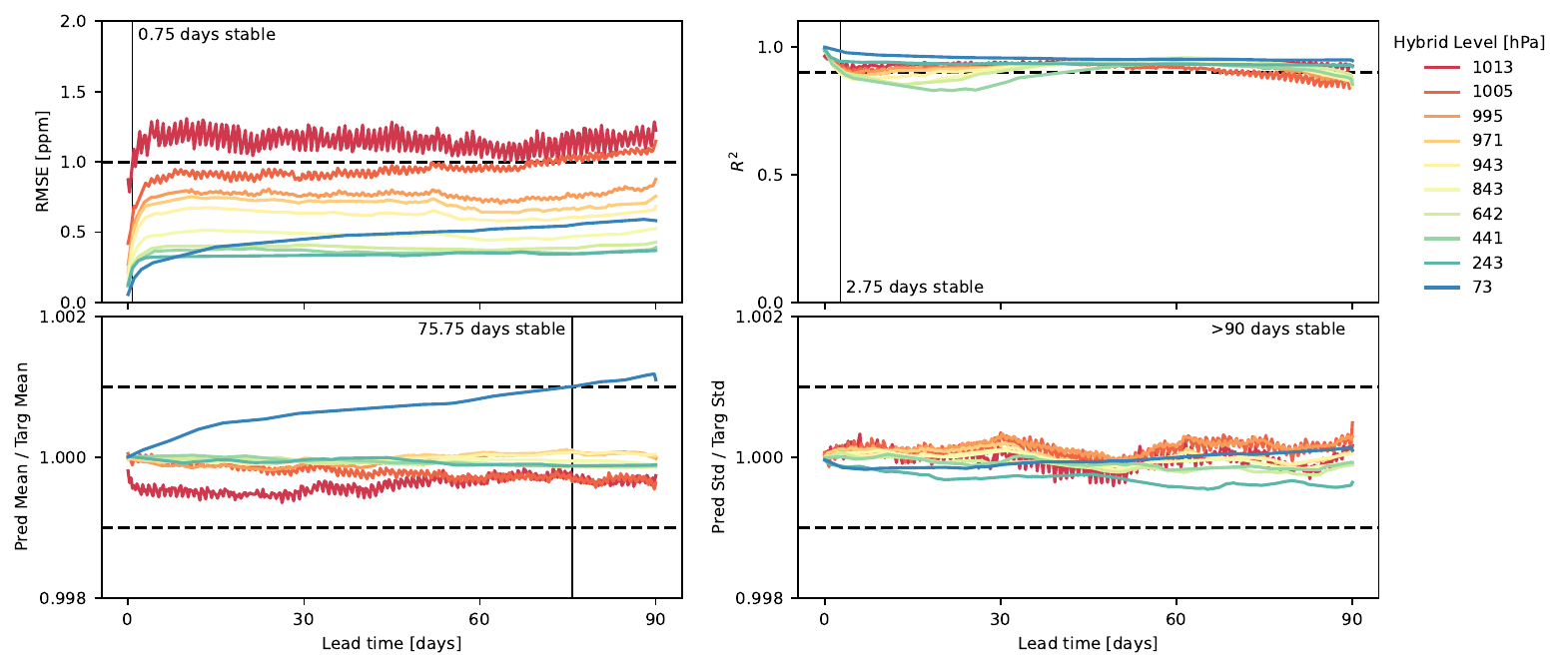}
    \caption{Same as fig.~\ref{fig:keymetrics_swintransformer} but for GraphCast.}
    \label{fig:keymetrics_graphcast}
\end{figure}

\begin{figure}
    \centering
    \includegraphics[width = \textwidth]{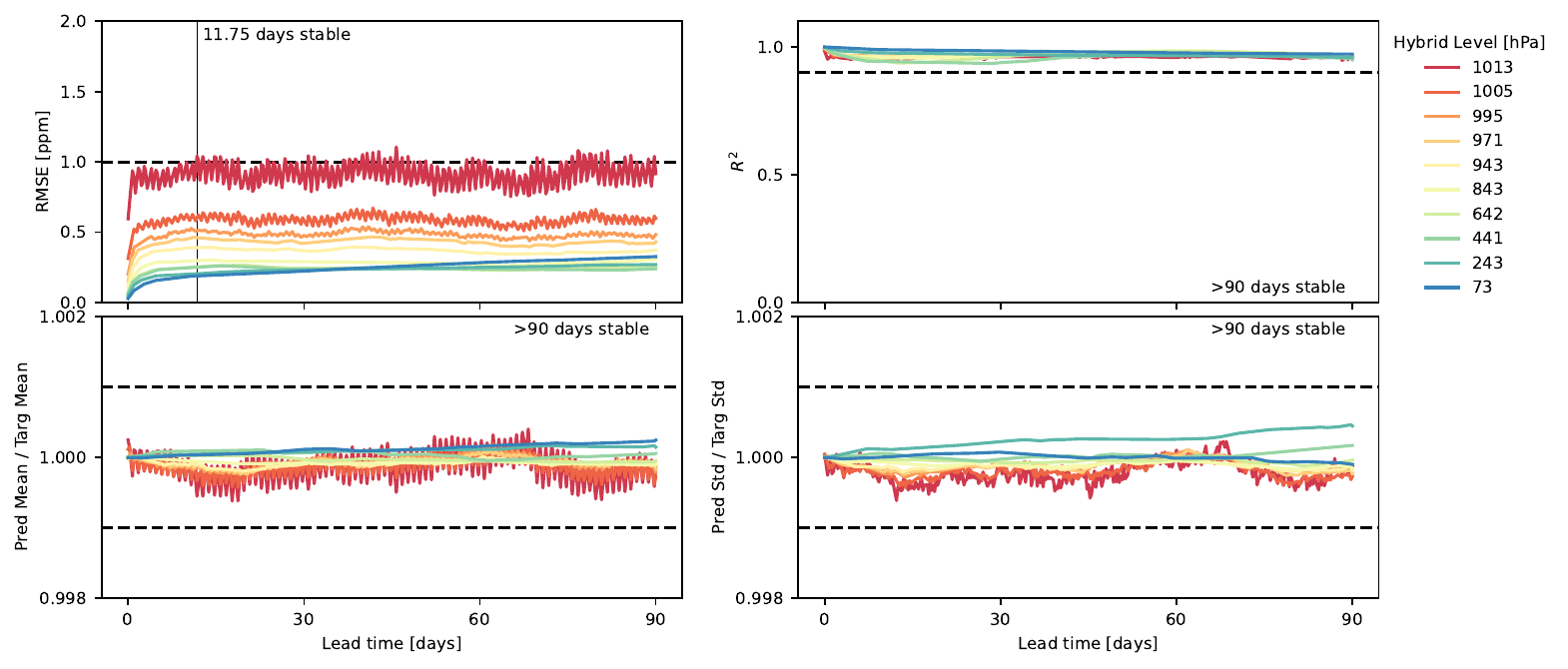}
    \caption{Same as fig.~\ref{fig:keymetrics_swintransformer} but for SFNO.}
    \label{fig:keymetrics_sfno}
\end{figure}

\begin{figure}
    \centering
    \includegraphics[width = \textwidth]{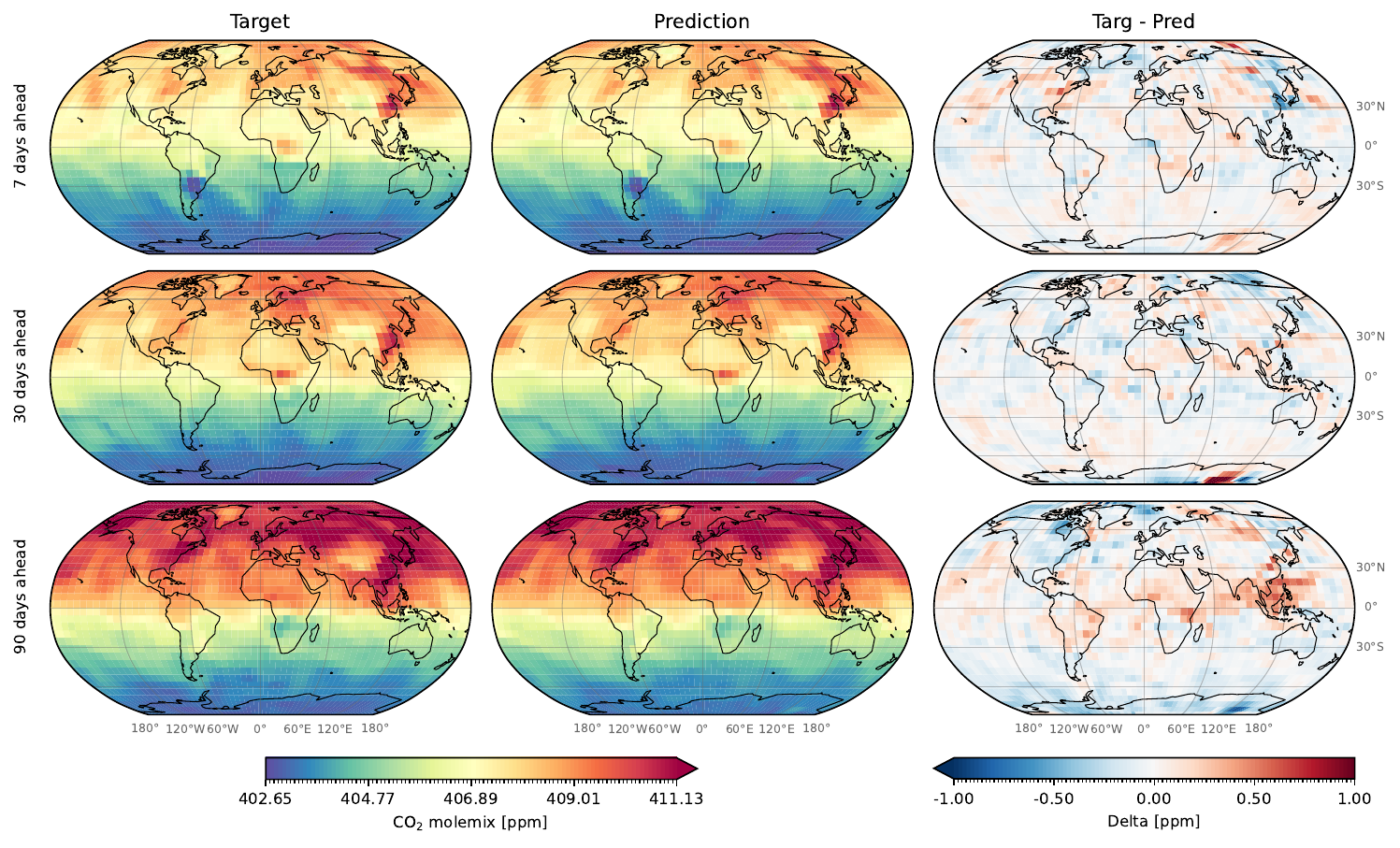}
    \caption{Same as fig.~\ref{fig:maps_swintransformer} but for UNet.}
    \label{fig:maps_unet}
\end{figure}

\begin{figure}
    \centering
    \includegraphics[width = \textwidth]{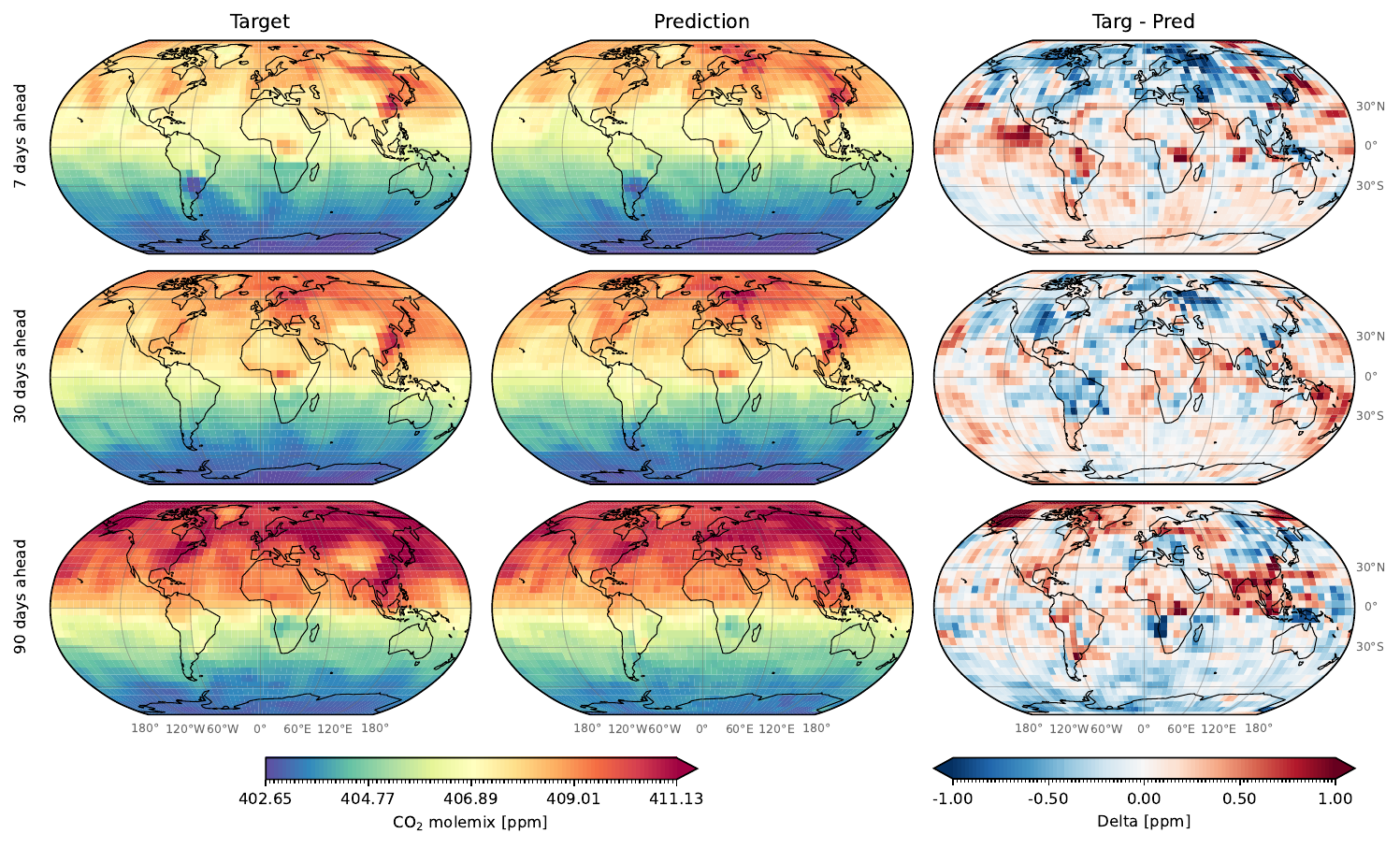}
    \caption{Same as fig.~\ref{fig:maps_swintransformer} but for GraphCast.}
    \label{fig:maps_graphcast}
\end{figure}

\begin{figure}
    \centering
    \includegraphics[width = \textwidth]{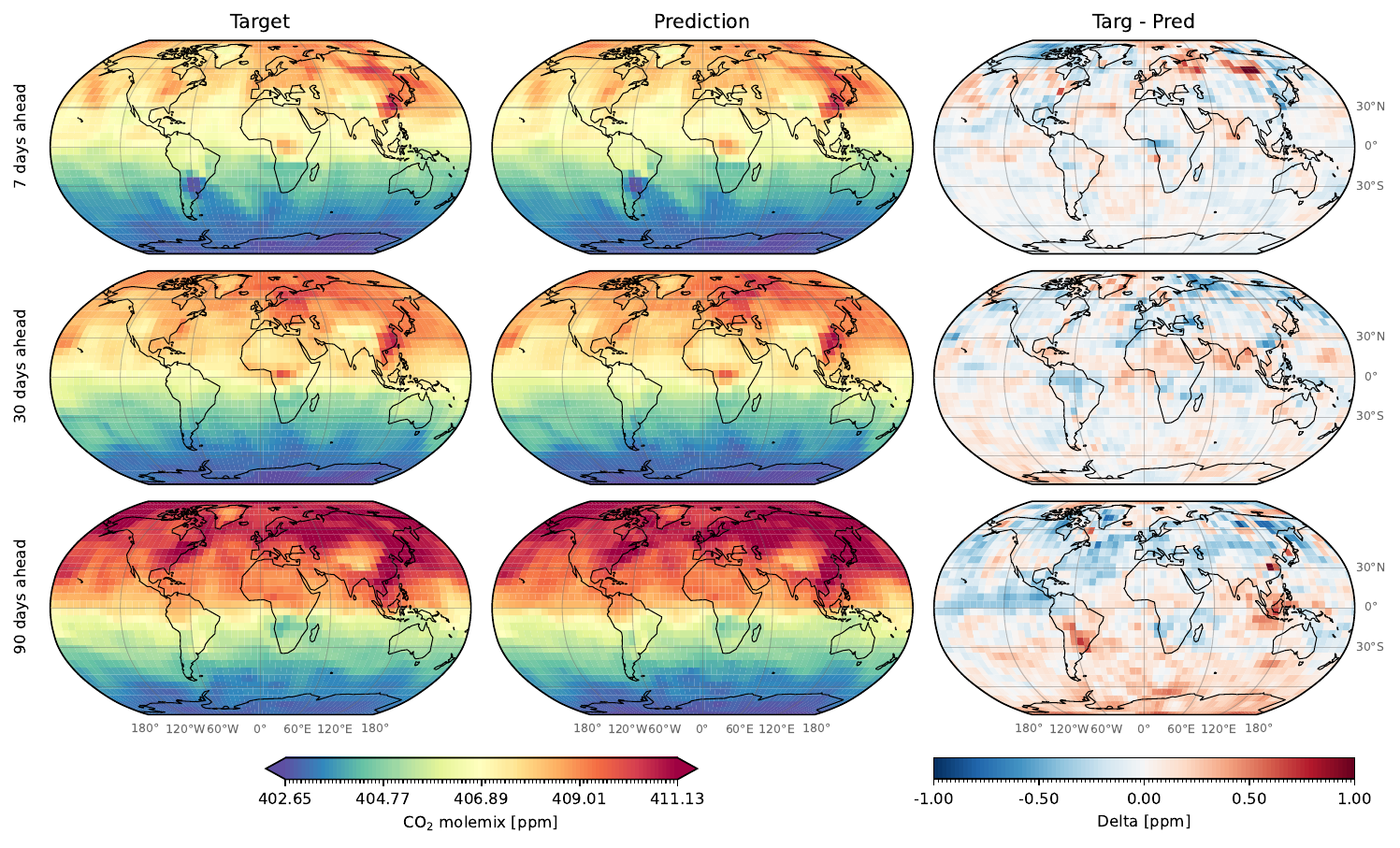}
    \caption{Same as fig.~\ref{fig:maps_swintransformer} but for SFNO.}
    \label{fig:maps_sfno}
\end{figure}

\begin{figure}
    \centering
    \includegraphics[width = \textwidth]{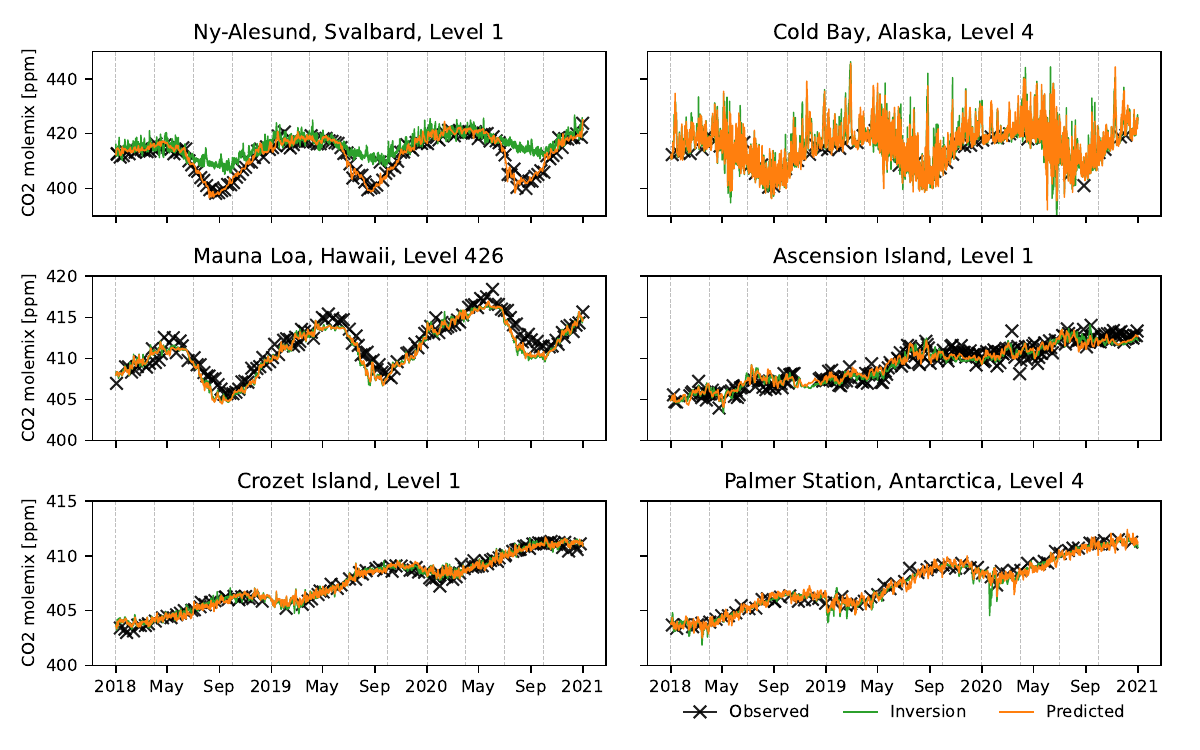}
    \caption{Same as fig.~\ref{fig:obspack_swintransformer} but for UNet.}
    \label{fig:obspack_unet}
\end{figure}

\begin{figure}
    \centering
    \includegraphics[width = \textwidth]{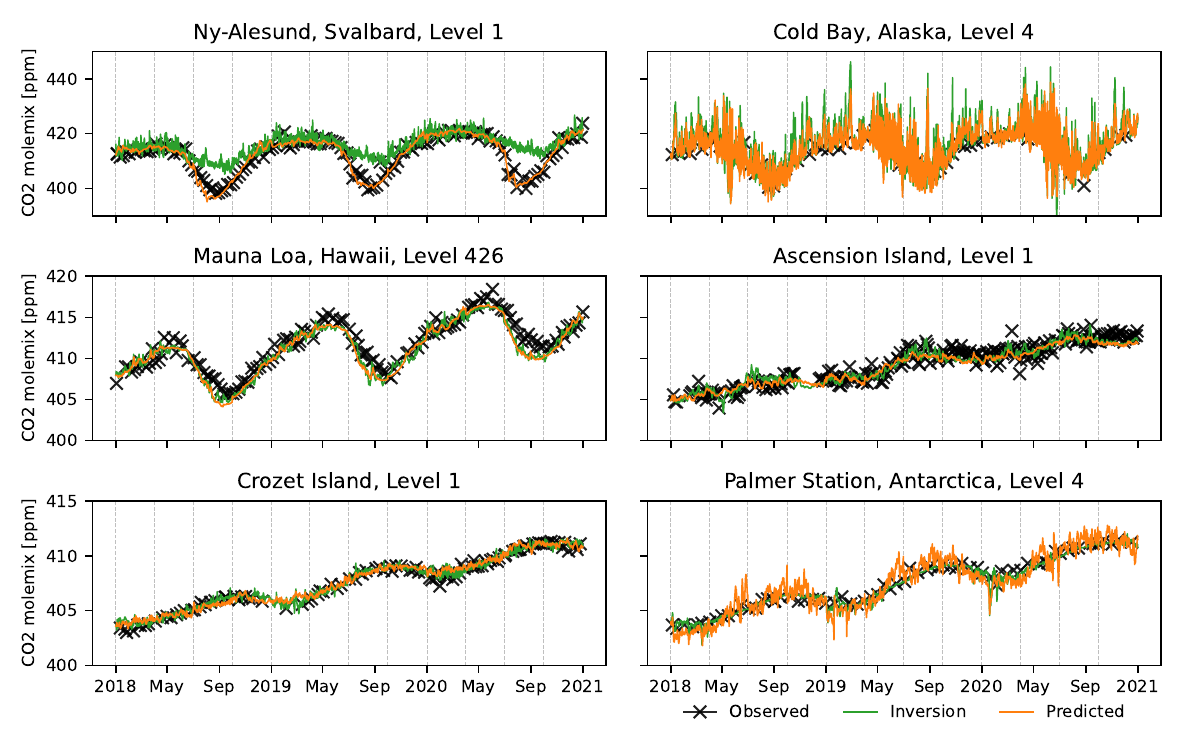}
    \caption{Same as fig.~\ref{fig:obspack_swintransformer} but for GraphCast.}
    \label{fig:obspack_graphcast}
\end{figure}

\begin{figure}
    \centering
    \includegraphics[width = \textwidth]{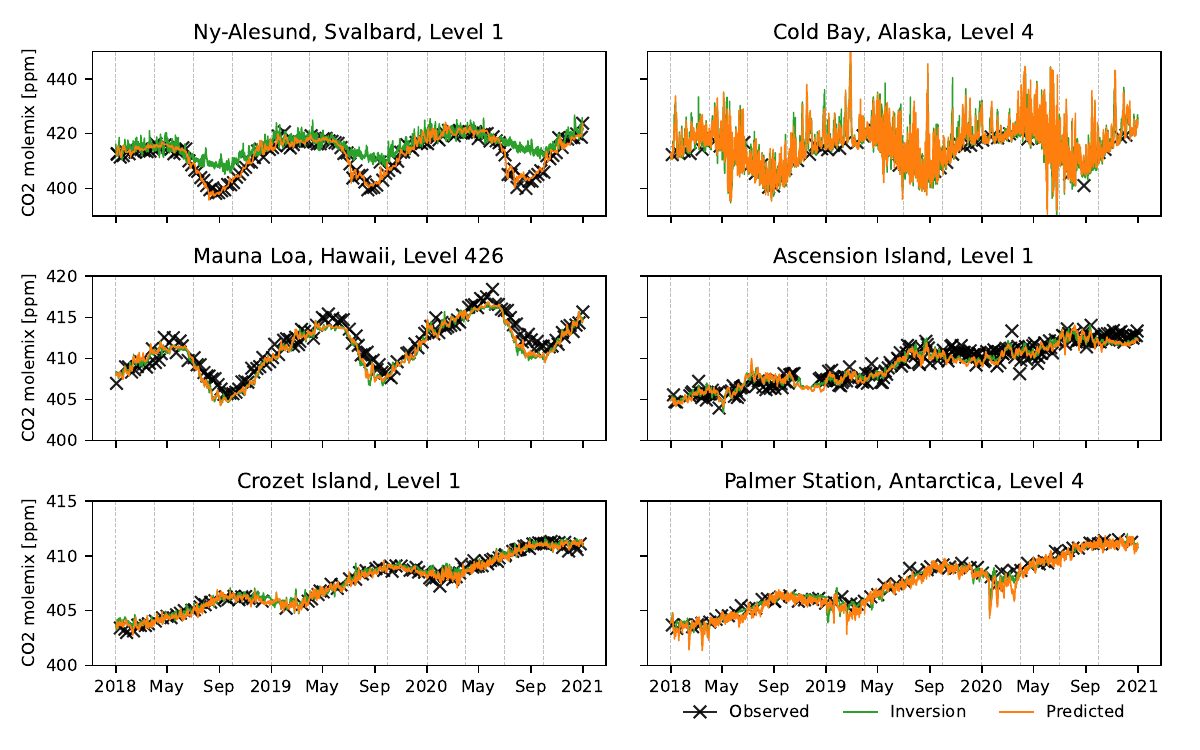}
    \caption{Same as fig.~\ref{fig:obspack_swintransformer} but for SFNO.}
    \label{fig:obspack_sfno}
\end{figure}

\begin{figure}
    \centering
    \includegraphics[width = \textwidth]{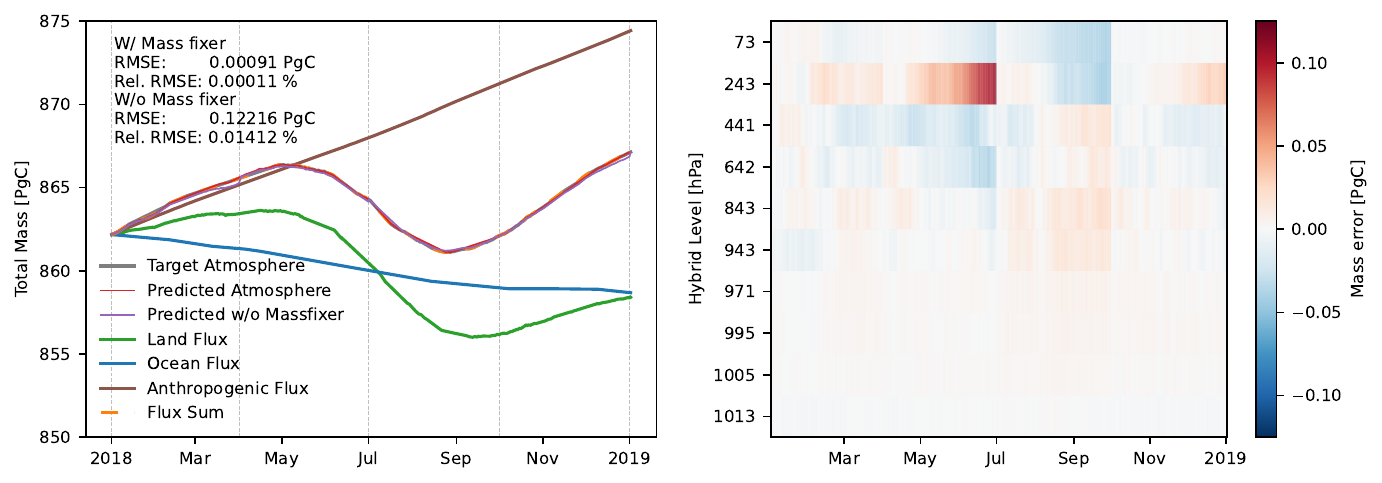}
    \caption{Same as fig.~\ref{fig:mass_swintransformer} but for UNet.}
    \label{fig:mass_unet}
\end{figure}

\begin{figure}
    \centering
    \includegraphics[width = \textwidth]{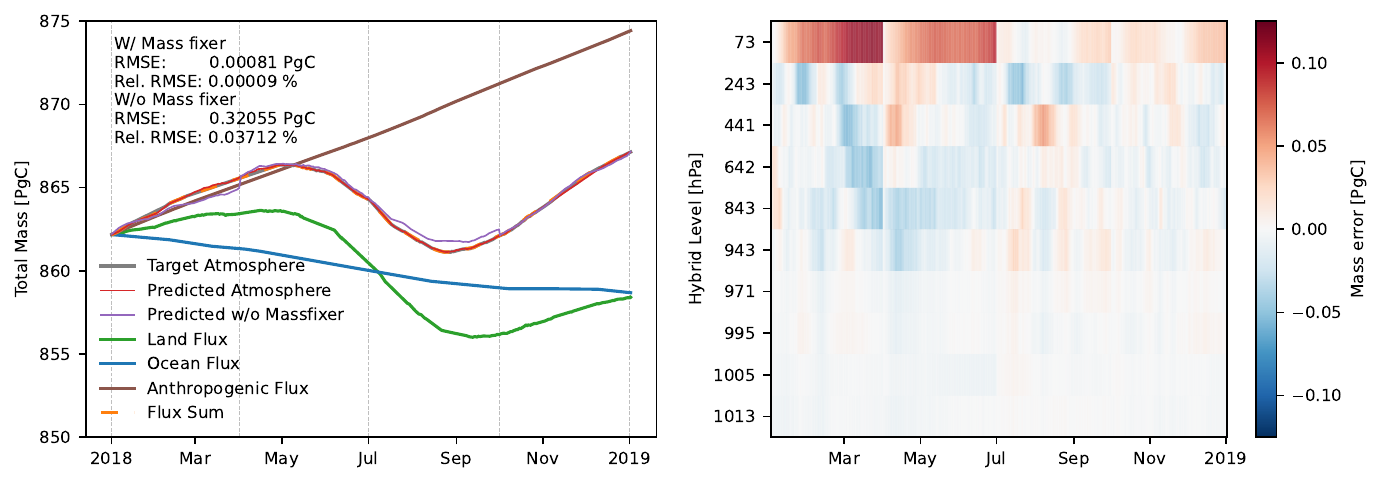}
    \caption{Same as fig.~\ref{fig:mass_swintransformer} but for GraphCast.}
    \label{fig:mass_graphcast}
\end{figure}

\begin{figure}
    \centering
    \includegraphics[width = \textwidth]{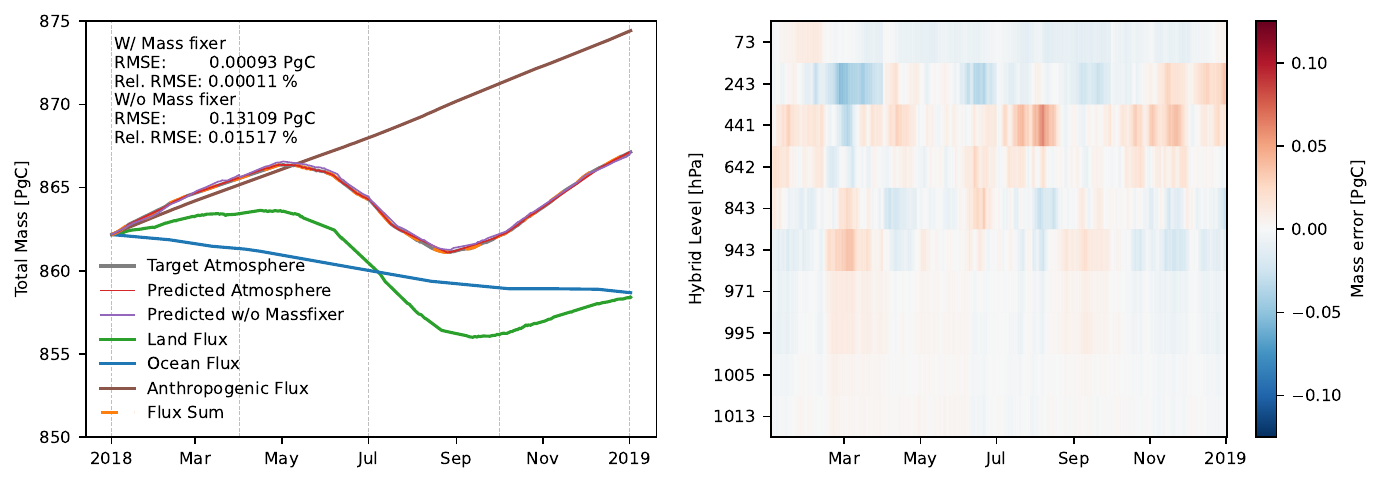}
    \caption{Same as fig.~\ref{fig:mass_swintransformer} but for SFNO.}
    \label{fig:mass_sfno}
\end{figure}

\clearpage

\section*{References}
\printbibliography[heading=none]
\fi 

\end{document}